\documentclass[sigconf,nonacm, screen]{acmart}
\usepackage{amsmath}
\usepackage{bm}
\usepackage{amsfonts}
\newcommand{\req}[1]{Eq.~(\ref{#1})}
\newcommand{\rfig}[1]{Fig.~\ref{#1}}
\newcommand{\rtab}[1]{Tab.~\ref{#1}}

\theoremstyle{plain}
\newtheorem{theorem}{Theorem}[section]

\theoremstyle{definition}
\newtheorem{definition}[theorem]{Definition}

\theoremstyle{remark}

\usepackage{graphicx}

\usepackage{bm}
\usepackage[labelformat=simple]{subcaption}

\usepackage{algorithm}
\usepackage[noend]{algpseudocode}

\usepackage{booktabs} 

\def\figdir{MinFigs/}

\AtBeginDocument{%
  }

\begin{document}

\title{Evaluating Time-Series Training Dataset through Lens of Spectrum in Deep State Space Models}
\author{Sekitoshi Kanai}

\email{sekitoshi.kanai@ntt.com}
\affiliation{%
  \institution{NTT}
  \state{Tokyo}
  \country{Japan}
}

\author{Yasutoshi Ida}
\affiliation{%
  \institution{NTT}
  \state{Tokyo}
  \country{Japan}
}

\author{Kazuki Adachi}
\affiliation{%
  \institution{NTT}
  \state{Tokyo}
  \country{Japan}
}

\author{Mihiro Uchida}
\affiliation{%
  \institution{NTT}
  \state{Tokyo}
  \country{Japan}
}

\author{Tsukasa Yoshida}
\affiliation{%
  \institution{NTT}
  \state{Tokyo}
  \country{Japan}
}

\author{Shin'ya Yamaguchi}
\affiliation{%
  \institution{NTT}
  \state{Tokyo}
  \country{Japan}
}

\renewcommand{\shortauthors}{Kanai et al.}

\begin{abstract}
  This study investigates a method to evaluate time-series datasets in terms of the performance of deep neural networks (DNNs) with state space models (deep SSMs) trained on the dataset.
  SSMs have attracted attention as components inside DNNs to address time-series data.
 Since deep SSMs have powerful representation capacities, training datasets play a crucial role in solving a new task.
 However, the effectiveness of training datasets cannot be known until deep SSMs are actually trained on them. 
  This can increase the cost of data collection for new tasks,
  as a trial-and-error process of data collection and time-consuming training are needed to achieve the necessary performance.
  To advance the practical use of deep SSMs, the metric of datasets to estimate the performance early in the training can be one key element.    
  To this end, we introduce the concept of data evaluation methods used in system identification.
  In system identification of linear dynamical systems, the effectiveness of datasets is evaluated by using the spectrum of input signals.
  We introduce this concept to deep SSMs, which are nonlinear dynamical systems.
  We propose the K-spectral metric, which is the sum of the top-K spectra of signals inside deep SSMs,
  by focusing on the fact that each layer of a deep SSM can be regarded as a linear dynamical system.
  Our experiments show that the K-spectral metric has a large absolute value of the correlation coefficient with the performance and can be used to evaluate the quality of training datasets.\looseness=-1
\end{abstract}


\maketitle

\section{Introduction}
\label{Intro}
Time-series data are ubiquitous in various fields~\cite{KDDTuto}, such as healthcare~\cite{zhang2023warpformer,harutyunyan2019multitask}, industrial IoT~\cite{matsubara2019dynamic,liu2020deep}, and finance~\cite{sezer2020financial}.
To analyze time-series data, machine learning methods continue to be studied and explored~\cite{esling2012time,liao2005clustering,sapankevych2009time}, and
recent deep neural network (DNN)-based methods have enabled us to analyze complicated time-series data~\cite{purushotham2018benchmarking,liu2020deep,sezer2020financial}.
Especially, DNNs with structured state space models (SSMs), e.g., S4~\cite{S4} and S5~\cite{S5}, have attracted much attention because they can address long-term dependencies~\cite{S4D, hippo}.
However, DNNs with SSMs (deep SSMs) require a large data sample size, which can be a bottleneck in practical data analysis applications.
Moreover, when encountering a new task, 
we do not know  whether a prepared dataset has sufficient information to solve the task.
We can determine whether the dataset is effective to solve the given task only after training.
Therefore, data scientists often need to collect data and train iteratively until a satisfactory performance is obtained.
In fact, MLOps has a feedback loop from the training process to the data engineering process~\cite{mlops}.
The cost of such iterative trial-and-error runs can be reduced
if we can estimate the performance of trained deep SSMs early in the training.

To evaluate the effectiveness of training datasets, one candidate metric is the data sample size.
\citet{rosenfeld2020a} have presented fitting a power law function to show the relation between the performance and data sample size~\cite{bahri2021explaining}.
\citet{mahmood2022much} have presented fitting more general functions.
Another candidate is validation loss at the first few epochs. 
However, since these approaches implicitly assume that the information about the tasks is uniformly distributed over data samples,
they do not necessarily evaluate the effectiveness of real-world training datasets precisely.
Real-world data can lack specific data samples due to bias in the data collection process.
For example, sensor data collection of running chemical plants 
is difficult to include all reachable states of the plant
because there are plant-friendly constraints, e.g., minimizing variability in product quality~\cite{rivera2003plant}.
Additionally, simply collecting and integrating all of data may have a negative affect on model training in some cases~\cite{roh2019survey}. 
Thus, methods need to be developed that can evaluate various training datasets including biased datasets.

\begin{figure}[tb]
    \centering
\includegraphics[width=\linewidth]{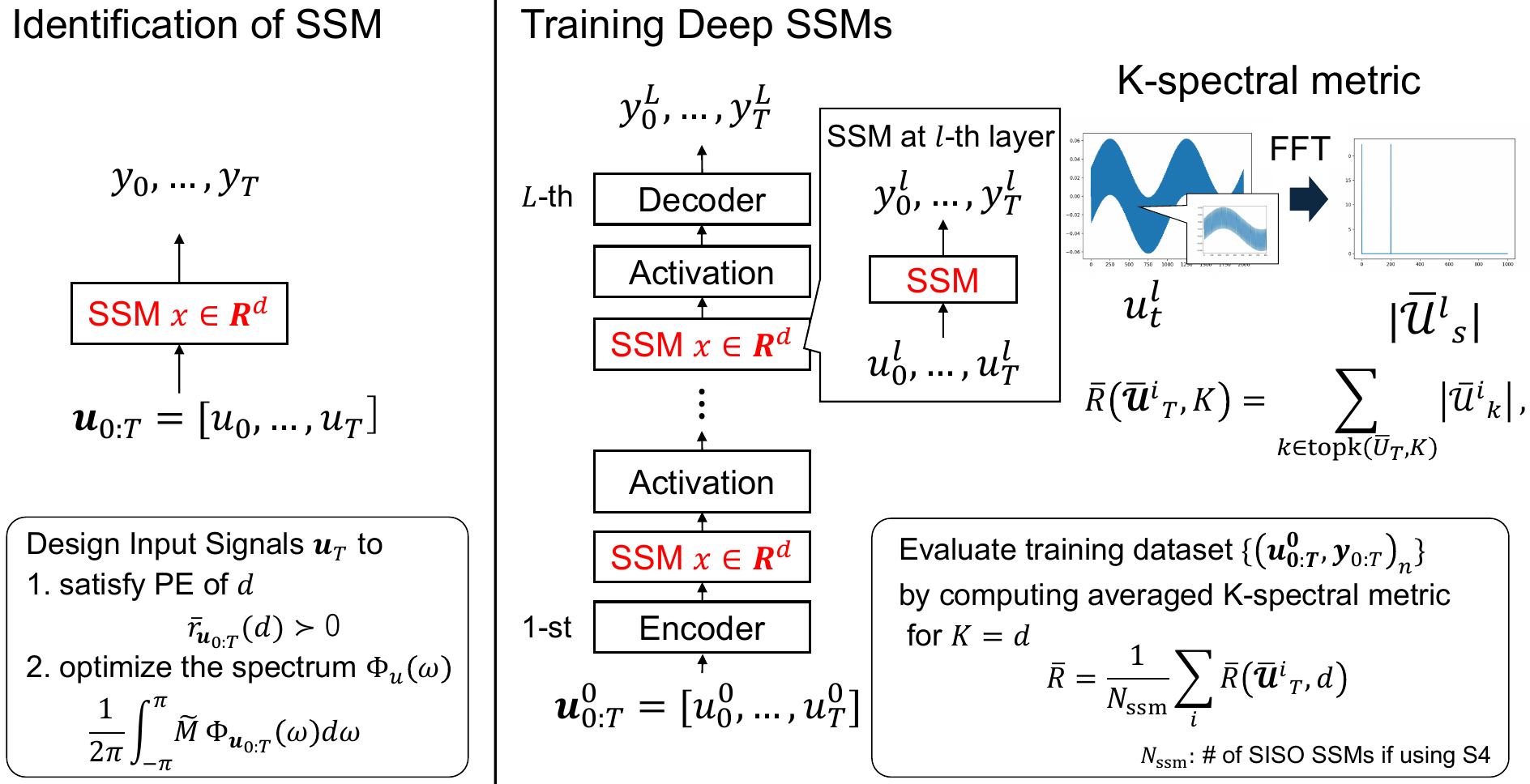}
\caption{Overview of evaluation of a training dataset by K-spectral metric (right) compared with input design of system identification (left). 
K-spectral metric is a sum of top-K magnitudes of frequency components $|\bar{U}^l_s|$ of $u^l_t$ applied to SSMs.}
    \label{Concept}
\end{figure}

In this paper, we propose a new metric called the K-spectral metric that correlates with the test performance of deep SSMs trained on various training datasets on the basis of the concept in linear system identification.
To evaluate the effectiveness of training datasets, we introduce the concepts of the \textit{optimal input design} and \textit{Persistence of Excitation (PE)} in system identification~\cite{ljung1999system}.
In system identification, we need to collect training datasets to build a model of a target physical system by applying input signals to the system and observed output signals.
The optimal input design explores the input signal to minimize the estimation errors of parameters.
The optimality is determined by the spectrum rather than the shape of input signals when the target system is a linear dynamical system.
Roughly speaking, the optimal input signals have a large magnitude on the sensitive frequency area of the system~\cite{mehra1974optimal,rojas2007robust,ljung1999system}.
If there is no a priori knowledge about target systems, PE becomes the metric of the informativeness of training data~\cite{ljung1999system}.
The PE condition corresponds to the  number of frequency components of input signals, 
and it should be large enough for identifying higher-order systems.
Though the spectrum is an important metric for linear systems as above, 
it is not obvious whether it is also a useful metric for training datasets of deep SSMs, which are nonlinear systems. 
Since deep SSMs have linear systems inside the model architecture, we investigate the following questions:\looseness=-1
{
\setlength{\leftmargini}{12pt}
\begin{itemize}
\item \textit{Do the frequency components of intermediate signals before SSMs represent the effectiveness of training data when used in DNNs?}
\item \textit{If so, how can we use them to evaluate training datasets?}
\end{itemize}}
To answer these questions, we empirically investigate the relationship between the performance and the frequency components.
Specifically, we evaluate the sum of top-K magnitudes of frequency components of intermediate signals that are applied to SSMs in deep SSMs on the basis of
the concept of optimal input design and PE (\rfig{Concept}).
We name this metric as the K-spectral metric
and experimentally show that it can evaluate the effectiveness of training datasets on
nonlinear system identification, classification, and forecasting problems of time-series data.
The main contributions of this paper are as follows:
{
\setlength{\leftmargini}{12pt}
\begin{itemize}
    \setlength{\topsep}{3pt}
    \item We propose K-spectral metric, a performance-correlated metric that can evaluate the effectiveness of training datasets.
    Its correlation coefficients are larger than those of the data sample size and validation loss when we compare training datasets 
    that lack data samples uniformly and biasedly.
    \item We empirically reveal that the spectrum of intermediate signals of deep SSMs (i.e., the K-spectral metric) is important for solving complex tasks
    although the whole model architecture is a nonlinear dynamical system.\looseness=-1
    \item Our experiments reveal that a flatter spectrum of the intermediate signals is required for deep SSMs to achieve good performance early in the training.
    After the middle of the training, the spectrum should be the specific shape needed for the problem.
\end{itemize} 
}
\section{Preliminaries}
\subsection{State Space Models in DNNs}
After \citet{hippo} have revealed that certain SSMs called HiPPO address 
the long-term dependency, several studies used SSMs in DNNs for time-series data~\cite{S4, S4D, S5, xiao2023imputation,goel2022s,miyazaki2023structured}.
S4~\cite{S4} initializes their parameters to satisfy the HiPPO framework and update them to learn time-series data.
Whereas S4 addresses the multiple inputs by using several single-input and single-output (SISO) SSMs,
S5~\cite{S5} extends S4 to address multiple inputs with the HiPPO initialization 
by using only one multiple-input and multiple-output (MIMO) SSM.
Let $u_t^{l}\in\mathbb{R}$, $y_t^{l}\in \mathbb{R}$, and $\bm{x}_t^{l}\in\mathbb{R}^{d}$ be an input, output, and state vector of the $l$-th layer at a discrete time-step $t$, respectively.
A linear time-invariant SISO SSM is written as:\looseness=-1 
\begin{align}
    \bm{x}^{l}_{t}&=\bm{A}^{l}\bm{x}^{l}_{t-1}+\bm{b}^{l}u^{l}_{t-1},\label{ssm}\\
    y^{l}_t&=\bm{c}^{l\mathsf{T}}\bm{x}^{l}_t+D^{l}u^{l}_t,
\end{align}
where $\bm{A}^l\!\in\!\mathbb{R}^{d\times d}$, $\bm{b}^l\!\in\!\mathbb{R}^{d\times 1}$, $\bm{c}^l\!\in\!\mathbb{R}^{d \times 1}$, and $D^l\!\in\!\mathbb{R}$ 
are parameters.
The input $u^{l}_t$ of the $l$-th layer is the output of the $(l-1)$-th layer after an activation function $\phi$:
$u^{l}_t=\phi(y^{l-1}_t)$.
$\phi(y^{l-1}_t)$ is generally a vector, and thus,
input can also be a vector: $\bm{u}^{l}_t\!=\![u^{l,1}_t,\dots,u^{l,d_{\mathrm{in}}}_t]^\mathsf{T}$.
S4 considers $d_{\mathrm{in}}$ SISO SSMs for one layer,
and we consider our metric for each element of the input vector independently
even when using S5. 

Since a SSM is one of the representations of a linear dynamical system, 
it can be written by another representation.
Let $q$ be shift operator as $qu_t=u_{t+1}$ and $q^{-1}u_t=u_{t-1}$.
A discrete SSM can be written as a discrete transfer function $G_{\bm{\theta}}(q)$:
\begin{align}
    y_t&=G_{\bm{\theta}}(q)u_t,\\
    G_{\bm{\theta}}(q)&=\bm{c}^{\mathsf{T}}(q \bm{I}-\bm{A})^{-1}\bm{b}+D.
\end{align}
where $\bm{\theta}$ denotes parameters of a system and $\bm{\theta}\!=\![\mathrm{vec}(\bm{A})^\mathsf{T},\bm{b}^{\mathsf{T}},\bm{c}^{\mathsf{T}},D]^\mathsf{T}$ in this case.
Similarly, a SSM can be also approximated by a finite impulse response model (FIR):
\begin{align}\textstyle
    y_t&\textstyle\approx Du_t+\bm{c}^{\mathsf{T}}\bm{b}u_{t-1}+\bm{c}^{\mathsf{T}}\bm{A}\bm{b}u_{t-2}+\dots+\bm{c}^{\mathsf{T}}\bm{A}^{d^\prime-1}\bm{b}u_{t-d^\prime}\\
&\textstyle=\sum_{i=1}^{d^\prime} \theta_i u_{t-i}.
\end{align}
If the SSM is stable, $\lim_{d^\prime\rightarrow \infty}\bm{c}^{\mathsf{T}}\bm{A}^{d^\prime}\bm{b}$ converges to zero.
Thus, a SSM can be written by the FIR with sufficient large $d^\prime$.
We use these representations to explain the optimal input design and PE simply. 
\subsection{Optimal Input Design}\label{OptInp}
System identification is a research area that builds mathematical models for dynamical systems
to control them~\cite{ljung1999system}. 
In system identification, we apply input signals $u_t$ to a physical system and observe output signals $y_t$.
Then, we estimate parameters $\bm{\theta}$ of the model from the datasets $\{(u_t,y_t)\}_{t=0}^{T-1}$. 
Input signals are designed to obtain accurate models. 
To minimize estimation errors of $\bm{\theta}$, 
we need to consider the spectrum of input signals rather than their waveforms~\cite{ljung1999system}.
The following objective function (A-optimality~\cite{aoki1970input,rojas2007robust,mithun2015d})
is often used to design input signals $\bm{u}_{0:T-1}=[u_0,\dots,u_{T-1}]$ as:\footnote{We consider an identification problem with additive white Gaussian noise $e_t$ as $y_t=G_{\bm{\theta}}(e^{j\omega})u_t+e_t$.}
\begin{align}
    \textstyle\bm{u}_{0:T-1}&\textstyle =\mathrm{arg}\max_{\bm{u}_{0:T-1}} \mathrm{tr}(\bm{M}),\label{a-op}\\
    \textstyle\bm{M}&\textstyle =\frac{1}{2\pi}\int_{-\pi}^\pi \tilde{\bm{M}}\Phi_{u_{0:T-1}}(\omega)d\omega,\label{FIM}\\
    \textstyle\tilde{\bm{M}}&\textstyle =\mathrm{Re}\left\{\frac{\partial G_{\bm{\theta}}(e^{j\omega})}{\partial \bm{\theta}}\left[\frac{\partial G_{\bm{\theta}}(e^{j\omega})}{\partial \bm{\theta}}\right]^H\right\}.
\end{align}
where $G_{\bm{\theta}}(e^{j\omega})$ is a rational transfer function parameterized by $\bm{\theta}$, which is a representation of the linear system in frequency domain.
$\omega$, $j$, and $H$ are angular frequency, imaginary unit, and Hermitian transpose, respectively. 
$\Phi_{u_{0:T-1}}$ is the spectral density:
\begin{align}\label{psd}\textstyle
    \Phi_{u_{0:T-1}}(\omega)=\lim_{T\rightarrow \infty}\frac{|\mathcal{U}_T(\omega)|^2}{T},
\end{align}
where $\mathcal{U}_T$ is Fourier transform of $\bm{u}_{0:T-1}$, which is a continuous-time input signal.
The objective function~(\req{a-op}) is derived through the Fisher information matrix (Appendix~\ref{FIMExApp}), which determines the bound of variance of unbiased estimators of $\bm{\theta}$~\cite{mehra1974optimal,rojas2007robust}.

\req{FIM} indicates that the optimal input signals should have the
spectral density depending on the target system in the frequency domain:
$\bm{\tilde{M}}$ can be regarded as the sensitivity of the frequency response to $\bm{\theta}$~\cite{ljung1999system}.
\subsection{Persistency of Excitation}\label{PESec}
PE is another important condition for input signals of system identification problems.
In system identification, $(u_t,y_t)$ should be \textit{informative}, i.e., the data allow discrimination between any two different models in a model set~\cite{ljung1999system}.
To achieve this, input signals $u_t$ should excite various oscillation modes of systems. 
The informative input is guaranteed by the metric called PE.
To grasp the concept of PE, we explain the PE condition by using a concrete example: identification of a FIR with white Gausssian noise $e_t$\looseness=-1
\begin{align}\label{fir}\textstyle
y_t=\sum_{i=1}^{d} \theta_i u_{t-i}+e_t.
\end{align}
We can estimate $\bm{\theta}$ by using $u_t$ and $y_t$ for $T$ time steps as:
\begin{align}
    \label{FIR}\textstyle
    \bm{\theta}&\textstyle=\left(\bm{U}_T^\mathsf{T}\bm{U}_T \right)^{-1}\bm{U}_T^\mathsf{T}\bm{y}_T,\\
    \textstyle \bm{y}_T&\textstyle=\left[\begin{smallmatrix}
        y_d\\
        y_{d+1}\\
        \vdots\\
        y_T
    \end{smallmatrix}\right], \bm{U}_T=\left[\begin{smallmatrix}
        u_{d-1}&u_{d-2}&\dots&u_0\\
        u_{d}&u_{d-1}&\dots&u_1\\
        \vdots&\vdots&\ddots&\vdots\\
        u_{T-1}&u_{T-2}&\dots&u_{T-d}\\
    \end{smallmatrix}.\right].\nonumber
\end{align}
To obtain the unique solution of \req{FIR}, the condition of $\mathrm{rank}\left(\bm{U}_T^\mathsf{T}\bm{U}_T \right)\!=\!d$ should be satisfied.
This condition corresponds to PE.
The PE is generally defined by using a covariance function:
\begin{definition}[\cite{ljung1999system}]
    Let $\bar{u}$ and $r_{u}(l)$ be an average and covariance function for input $u_t$ as
 $\bar{u}\!=\!\lim_{T\!\rightarrow\!\infty}\frac{1}{T}\!\sum_{t=0}^{T-1}u_t$ and $r_{u}(l)\!=\!\lim_{T\!\rightarrow\!\infty}\frac{1}{T}\!\sum_{t=0}^{T-1}(u_{t+l}-\bar{u})(u_t-\bar{u})$, respectively.
The input $u_t$ is persistently exciting of order $d$ if we have
$\bar{r}_{u}(d)\succ 0$ where
\begin{align}\textstyle
    \bar{r}_{u}(d)\!=\!\left[\begin{smallmatrix}
        r_{u}(0)&r_{u}(1)&\dots&r_{u}(d-1)\\
        r_{u}(1)&r_{u}(0)&\dots&r_{u}(d-2)\\
        \vdots&\vdots&\ddots&\vdots\\
        r_{u}(d-1)&r_{u}(d-2)&\dots&r_{u}(0)\\
    \end{smallmatrix}\right].
\end{align}
\end{definition}
For instance, one sinusoidal signal satisfies PE of order two,
and a sum of $m$ sinusoidal signals satisfies PE of order 2$m$.
White noise has all frequency components and satisfies PE of the infinite order: 
it satisfies $\bar{r}_u(d)\succ 0$, $\forall d$. 
PE also indicates that the spectral information of input signals can be the metric for evaluating datasets.
\subsection{Related Work}\label{RW}
\textbf{Deep SSM}.
\citet{hippo} have presented an SSM-based architecture called HiPPO.
Since HiPPO captures the dynamics of coefficients for the polynomial series restoring the original time transition function,
HiPPO can memorize the information of the original function.
After the HiPPO framework has been presented, several studies have presented deep SSMs~\cite{S4,S4D,S5}.
S4~\cite{S4} outperforms recurrent neural networks and Transformer variants in terms of time-series forecasting~\cite{S4}, anomaly detection~\cite{xiao2023imputation}, audio generation~\cite{goel2022s} and long-form speech recognition~\cite{miyazaki2023structured}.
\citet{S5} have extended S4 to address multiple inputs by one MIMO SSM and call this method S5.

\textbf{Evaluation of training dataset}.
\citet{roh2019survey} identified data evaluation as a future research challenge in data collection for machine learning: 
how to evaluate whether the right data was collected with sufficient quantity is an open question. 
The relationship between the dataset size and performance can fit a power law function~\cite{rosenfeld2020a,bahri2021explaining} and
more general functions~\cite{mahmood2022much}.
However, these approaches implicitly assume that the information for the tasks is uniformly distributed over data samples and cannot evaluate the effectiveness of
biased training datasets such that specific data samples are not obtained.
\citet{gupta2021data} have presented a toolkit for assessing various qualities of training datasets,
such as class overlap.
Since building the toolkit using several metrics is out of our research scope,
we do not compare our metric with this toolkit.
While \citet{sheng2008get} investigates labeling quality and propose repeated labeling,
we focus on input data points rather than target labels.
\looseness=-1

\textbf{PE in deep learning}.
Some studies have applied the concept of PE to deep learning~\cite{nar2019persistency, sridhar2022improving, zheng2017relationship, lekang2022sufficient} for various purposes.
\citet{nar2019persistency} and \citet{sridhar2022improving} have introduced the concept of PE in the dynamics of gradient descent, 
and \citet{lekang2022sufficient} have investigated the PE condition for the rectified linear unit (ReLU) activation.
To the best of our knowledge, this is the first study to investigate the relation between the performance
of the deep SSMs with the spectrum of intermediate signals.\looseness=-1
\section{Proposed Metrics}
Deep SSMs require a larger data sample size than traditional models, e.g., ARIMA for time-series data,
and the necessary quality of training datasets is not known. 
We consider the following problem: Let $\bm{u}^0_{0:T-1}\!=\![\bm{u}^0_0,\dots,\bm{u}^0_{T-1}]$ and $\bm{y}_{0:T-1}\!=\![y_0,\dots,y_{T-1}]$ be input data point and the target output, respectively. 
An $L$-layer deep SSM is trained on training dataset $\{(\bm{u}^0_{0:T-1},\bm{y}_{0:T-1})_n\}_{n=1}^N$ by the objective function $\frac{1}{N}\sum_n^N\mathcal{L}(\bm{y}_{0:T-1}^L,\bm{y}_{0:T-1})$ where $\bm{y}_{0:T-1}^L$
is output of the deep SSM.
\textit{Can the training dataset be evaluated with respect to the performance of deep SSMs?}

As discussed in Sections~\ref{OptInp} and \ref{PESec}, 
we are able to compare the given training datasets (input and output signals)
for system identification of linear systems in terms of optimality and PE before the parameter estimation. 
However, deep SSMs are generally nonlinear dynamical systems because $\phi$ contains nonlinear functions.
This makes their performance difficult to predict in advance of training.
Even so, linear dynamics of SSMs in the models are dominant in the dynamics of S4 and S5
because the state transition of each layer is linear computations.
Thus, we hypothesize that the spectrum of the input signal for each intermediate SSM
is related to the training performance like the optimal input design 
and the PE condition.

However, the optimal input design and PE condition are difficult to apply to deep SSMs directly.
Since we have no a priori knowledge of the target SSMs, we do not know $\tilde{\bm{M}}$ and cannot use the objective of optimal input design \req{a-op}.
Regarding PE, the computation of $\mathrm{rank}(\bar{r}_{u^l}(d))$ is numerically unstable and incurs high computation costs.
Thus, we consider the alternative computation by using the magnitudes of frequency components.

This section presents our proposed metric and explains why it is expected to correlate with performance through its relationship with optimal input design and PE.
\begin{figure*}[tb]
    \centering
    \subfloat[][$R(\bar{\bm{\mathcal{U}}}_T,K)$ for $K=12$]{\includegraphics[width=0.2\linewidth]{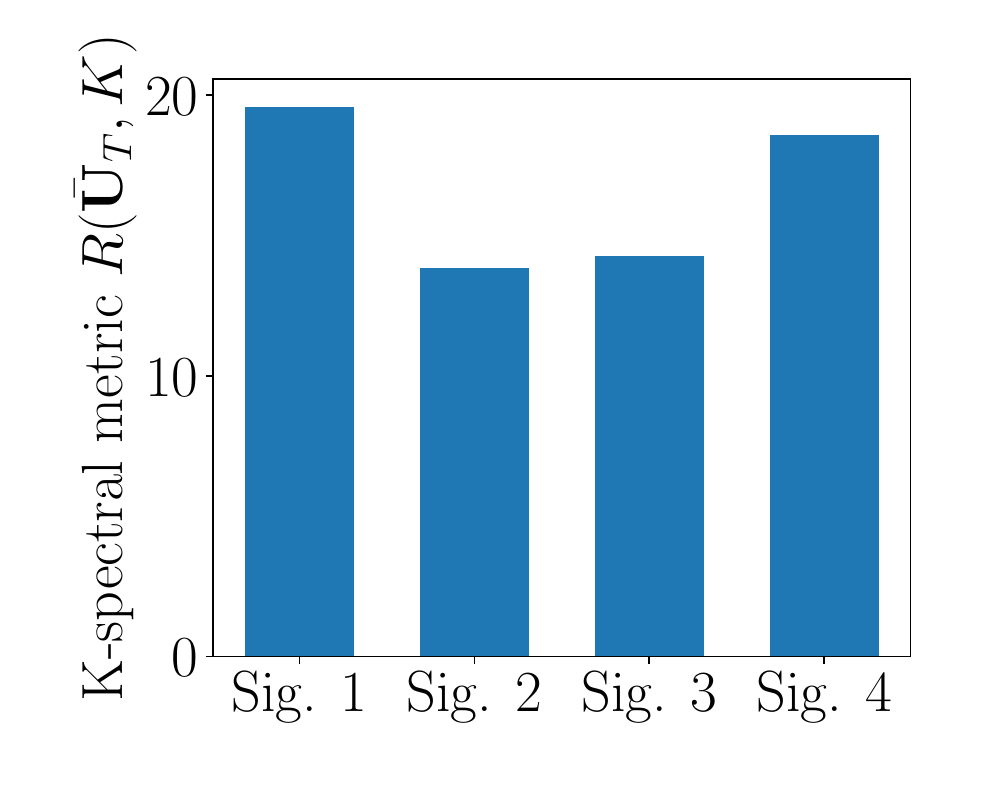}\label{Met}}
    \subfloat[][Signal 1 ($i\!=\!1$)]{\includegraphics[width=0.2\linewidth]{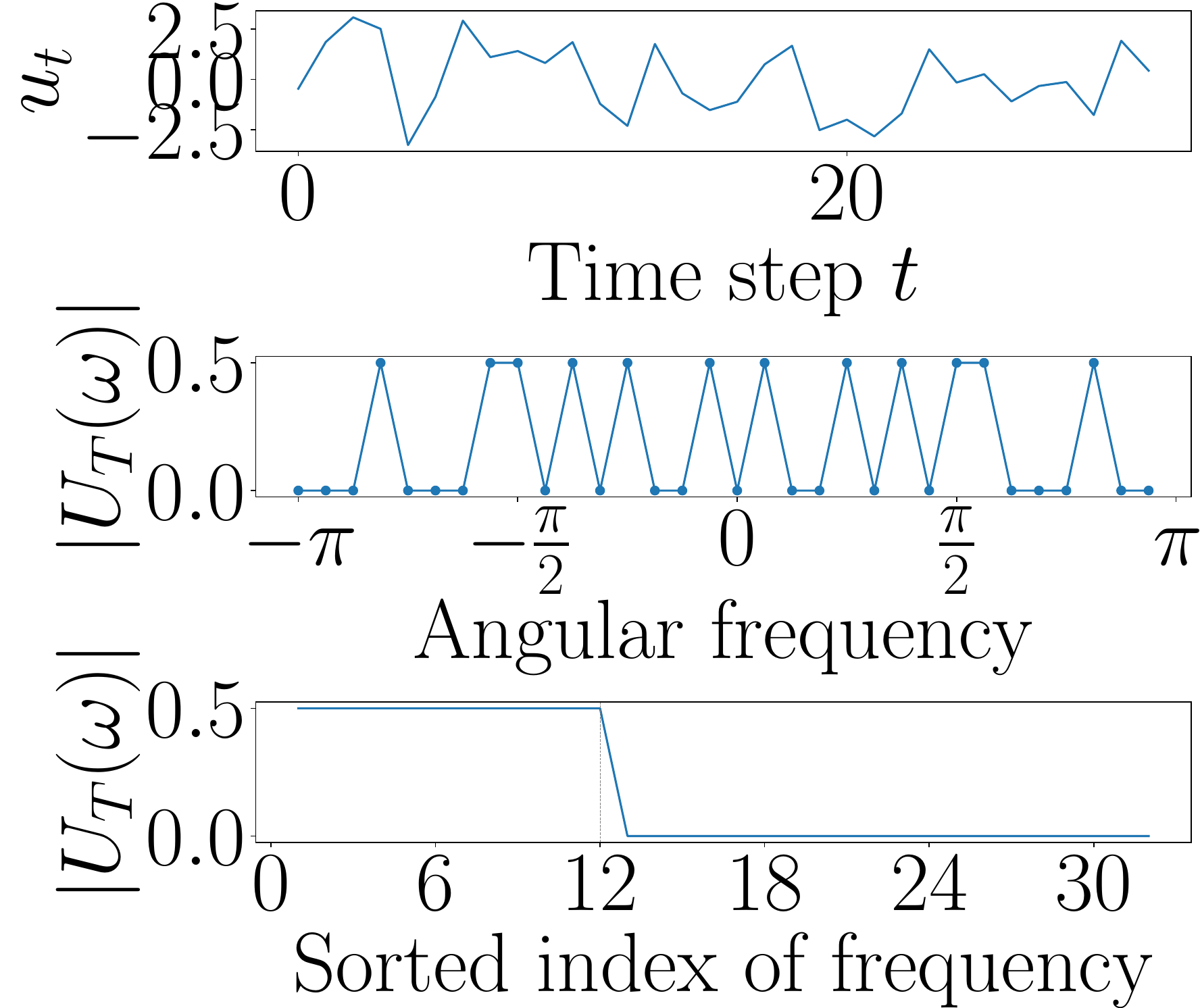}\label{Sig1}}
    \subfloat[][Signal 2 ($i\!=\!2$)]{\includegraphics[width=0.2\linewidth]{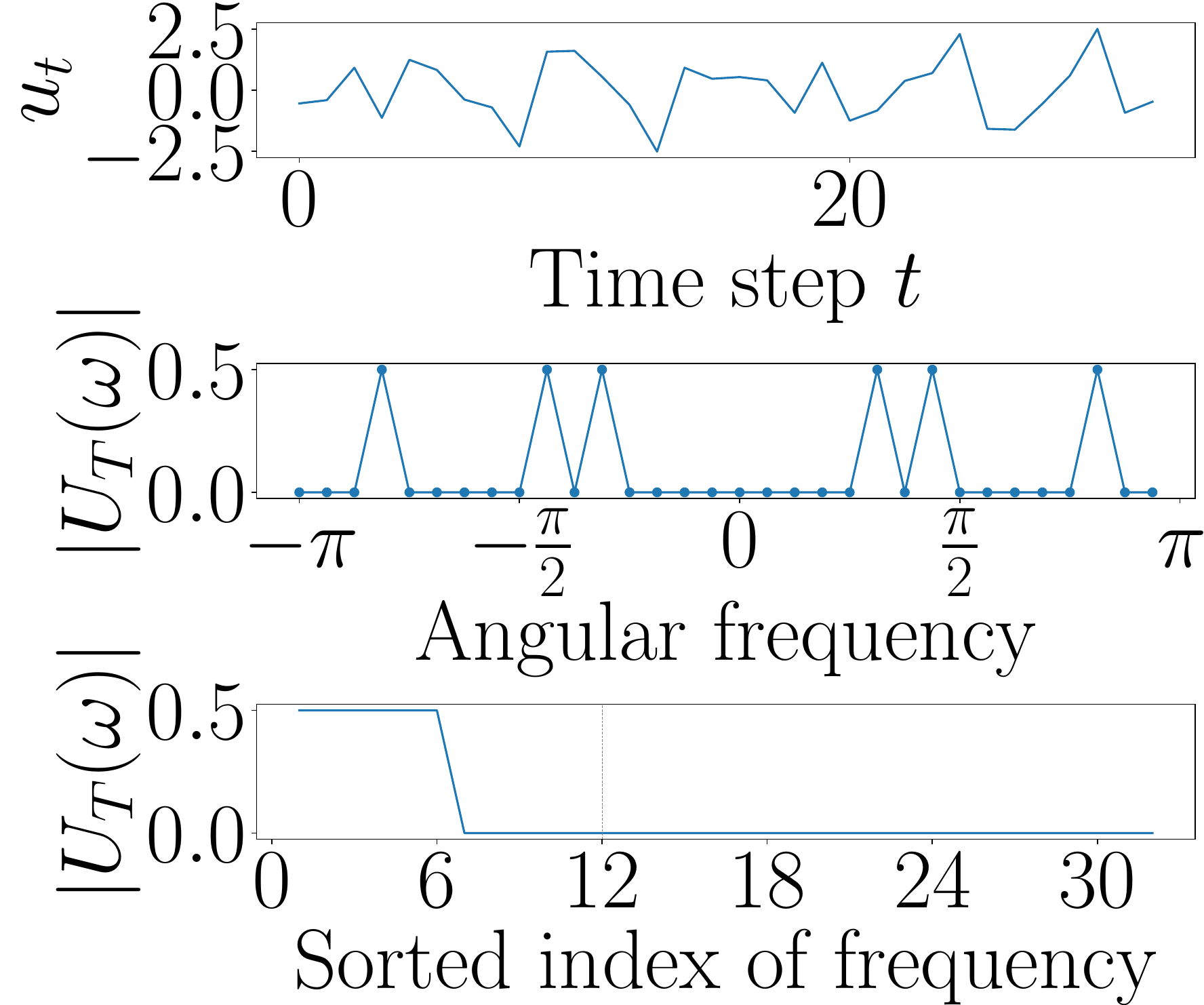}\label{Sig2}}
    \subfloat[][Signal 3 ($i\!=\!3$)]{\includegraphics[width=0.2\linewidth]{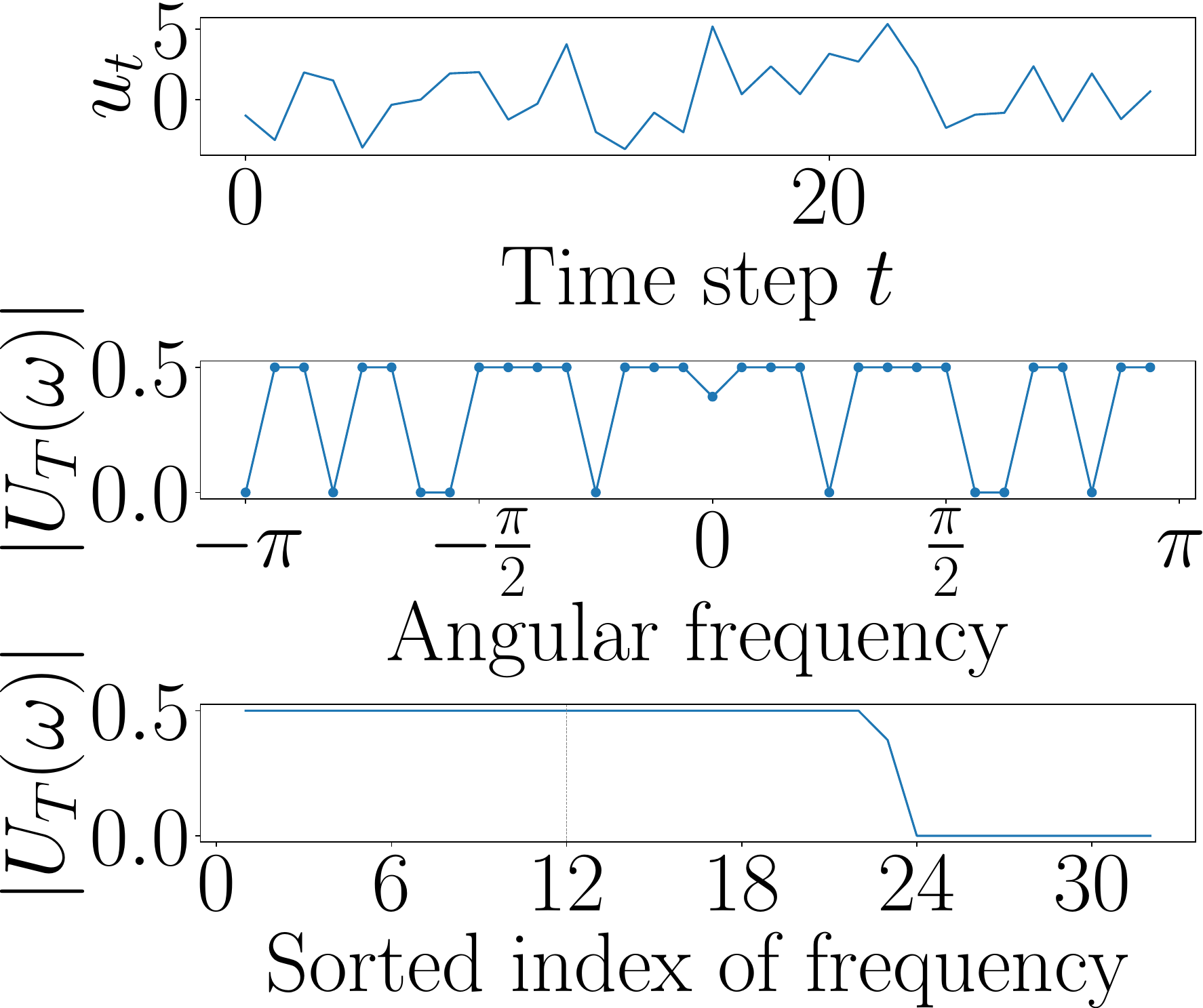}\label{Sig3}}
    \subfloat[][Signal 4 ($i\!=\!4$)]{\includegraphics[width=0.2\linewidth]{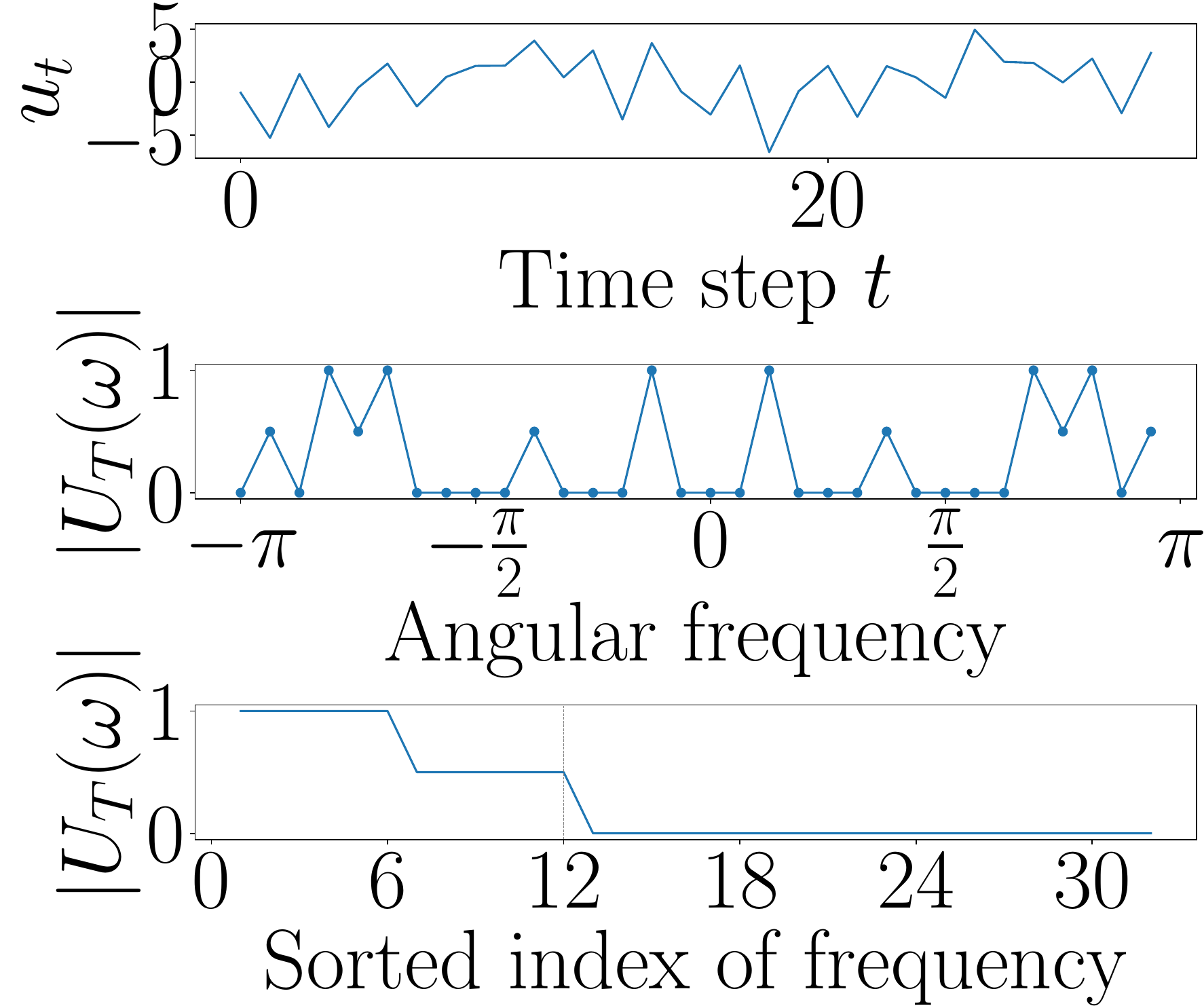}\label{Sig4}}
    \caption{K-spectral metrics for four multiple sinusoidal signals $u^i_t=\sum_s c^i_s \mathrm{sin}(2\pi st/T)$ for four settings $i=1,\dots,4$.
    \subref{Met} plots K-spectral metrics for these signals. 
    \subref{Sig1}-\subref{Sig4} are the waveforms (top), $|\mathcal{U}_T|$ by FFT (middle), and sorted $|\mathcal{U}_T|$ of signals (bottom). 
    When a signal has the flat spectrum for $K=12$ points, our metric achieves the highest.}
    \label{OurEx}
\end{figure*}
\subsection{K-spectral Metric}
We propose the K-spectral metric $R(\bar{\bm{\mathcal{U}}}_T,K)$, which is the sum of the top-K magnitudes of FFT of normalized input signals:
\begin{align}\label{prop}
    R(\bar{\bm{\mathcal{U}}}_T,K)&=\sum_{k\in \mathrm{topk}(\bar{\bm{\mathcal{U}}}_T,K)}|\bar{\mathcal{U}}_k|,
\end{align}
where  $\mathrm{topk}(S,K)$ is the function that outputs the index set that has indices of the largest $K$ elements for $S$.
$\bar{\bm{\mathcal{U}}}_T=[|\bar{\mathcal{U}}_0|,\dots,|\bar{\mathcal{U}}_{T-1}|]^\mathsf{T}$
is the magnitudes of the frequency components as
\begin{align}
    \bar{\mathcal{U}}_s\!=\!\sum_{t=0}^{T-1}\bar{u}_t e^{-j\omega t},~~\mathrm{for}~\omega\!=\!\frac{2\pi s}{T},~s=0,\cdots,T\!-\!1.
\end{align}
$\bar{u}_t$ is the normalized intermediate signal $\bar{u}_t\!=\!u_t/||\bm{u}_{0:T-1}||_2$
where $\bm{u}_{0:T-1}\!=\![u_0,\dots,u_{T-1}]^\mathsf{T}$.
Note that we omit the superscript of $l$ for simplicity.
Since a deep SSM can have multiple SSM layers and $d_{\mathrm{in}}$ intermediate signals in one layer, 
we average the K-spectral metric over all intermediate signals as:
\begin{align}
    \bar{R}=\frac{1}{N_{\mathrm{ssm}}}\sum_{i=1}^{N_{\mathrm{ssm}}}R(\bar{\bm{\mathcal{U}}}^i_T,K),
\end{align}
where $N_{\mathrm{ssm}}$ is the number of input signals for SSMs in a deep SSMs.
If $l_{\mathrm{ssm}}$ is the number of SSM layers,
$N_{\mathrm{ssm}}=d_{\mathrm{in}}l_{\mathrm{ssm}}$.
The hyperparameter $K$ is set to $d$ as explained in Section~\ref{RPESec}.
If $\bar{R}$ correlates with the test performance, 
we can use $\bar{R}$ as the metric of training datasets 
to estimate whether a training dataset has useful information for the task.
$\bar{R}$ is expected to correlate with performance because of the following two relationships.
\subsubsection{Relationship with Optimal Input Design}\label{RKOptIntSec}
The optimal input design suggests that
the spectrum $\Phi_u(\omega)$ for continuous-time signals determines
the optimality of input signals, i.e., training datasets, to estimate parameters as Eqs.~(\ref{a-op}) and (\ref{FIM}).
On the other hand, the K-spectral metric is based on the spectral density $|\bar{\mathcal{U}}_k|$ for discrete-time signals
because SSMs inside DNNs are always discrete-time systems.
Although the spectrum for continuous-time signals and the spectrum for discrete-time signals are different, 
they evaluate the effectiveness of training datasets through the magnitudes of the frequency domain rather than its phase, i.e., specific waveforms.

The spectrum of \req{FIM} is weighted by $\tilde{M}$
because system identification often uses a priori knowledge of the target system.
On the other hand, the K-spectral metric (\req{prop}) computes the sum of the spectral density of $K$ points without weighting
because we have no a priori knowledge.
Thus, a high K-spectral metric indicates that the input signals are optimal for estimating SSMs that have uniformly broad sensitivity $\tilde{\bm{M}}$ across frequency domains.
In other words, 
if the SSMs should have a uniform sensitivity to solve a task,
the K-spectral metric should become high, i.e., the K-spectral metric positively correlates with the performance. 
On the other hand, if the SSMs should have a non-uniform sensitivity after training,
the K-spectral should be low, i.e., the K-spectral metric negatively correlates with the performance.
While correlation coefficients can be both positive and negative,
our method can be used to estimate the performance because 
we observed that the absolute values of correlation coefficients are high enough in Section~\ref{experiments}.
\subsubsection{Relationship with PE}\label{RPESec}
A sinusoidal signal is persistently exciting of order two, and the sum of $m$ sinusoidal signals 
satisfies PE of $2m$. Conversely, if the input signal contains $m$ sinusoidal signals (i.e., spectrum is nonzero at $2m$ points) 
this signal satisfies PE of order $2m$. 
In other words, PE of higher order corresponds to more frequency components in the inputs.
If $|\bar{\mathcal{U}}_k|\!>\!0$ for $K$ points in frequency domain, 
input signals have $K$ frequency components and 
are expected to satisfy PE of order $K$.
We set $K$ to $d$ since SISO SSMs with $d$ states require input signals satisfying the PE of at least order $d$.
$|\bar{\mathcal{U}}_s|$ of the real signal is symmetric for $s$, but we use all of them.
The K-spectral metric has the following property:
\begin{theorem}\label{thm}
    $R(\bar{\bm{\mathcal{\mathcal{U}}}}_T,K)$ in \req{prop} achieves the maximum value
    $R^*_K=\max_{\bar{\bm{\mathcal{U}}}_T}R(\bar{\bm{\mathcal{U}}}_T,K)$ if and only if
        $|\bar{\mathcal{U}}_i|=|\bar{\mathcal{U}}_j|$ for all $i,j\in \mathrm{topk}(\bar{\bm{\mathcal{U}}}_T,K)$,
    and $|\bar{\mathcal{U}}_i|=0$ for $i \notin \mathrm{topk}(\bar{\bm{\mathcal{U}}}_T,K)$.
\end{theorem}
The proof is provided in Appendix~\ref{ProofApp}.
This theorem indicates that the K-spectral metric has the maximum value if
the magnitudes at $K$ points in the frequency domain have the same values and the others are zero.
This implies that if input signals are persistently exciting of higher order,
our metric $R(\bar{\bm{\mathcal{U}}}_T,K)$ also becomes higher for sufficiently large $K$. 
Since the K-spectral metric measures the informativeness of input signals
without the difficulty of rank computation, 
we use it for evaluating the training dataset quality.

\rfig{OurEx} shows how K-spectral metric of $K\!=\!12$ works for four 
multiple sinusoidal signals: $u^{i}_t\!=\!\sum_s c^{i}_s \mathrm{sin}(2\pi st/T+\psi_s)$
for $i\!=\!1,\dots,4$. $\psi_s$ is randomly selected phase shift.
We set $c^{i}_s=1$ for six randomly selected points of $s$ for Signal~1 ($i\!=\!1$),
three randomly selected points of $s$ for Signal~2, and twelve randomly selected points of $s$ for Signal~3.
For Signal~4, we set $c_s^4=1$ for randomly selected three points and $c_s^{4}=0.5$ for randomly selected three points of $s$.
We can see that Signal~1 achieves the highest K-spectral metric and the order is Signal~1 $>$ Signal~4 $>$ Signal~3 $>$ Signal~2.
This result follows Theorem~\ref{thm}: K-spectral metric becomes the largest when the signal has a uniform spectrum for $K$ points.
Signal~1 satisfies the PE condition of $K\!=\!12$ since one sinusoidal signal has two of the PE condition.
Since Signal~2 does not satisfy the PE condition of $K\!=\!12$,
its K-spectral metric has the lowest value.
When the signal has PE of higher order than $K$ (Signal~3), our metric becomes small. 
This characteristic is different from the PE condition.
As explained in Section~\ref{RKOptIntSec},
the K-spectral metric is uesd to evaluate the intermediate signals in terms of the optimality
once the input signal satisfies the PE condition.

\subsection{Implementation of K-spectral Metric}\label{Impl}
Since deep SSMs generally contain learnable layers before SSMs,
the learned layers affect the spectrum of the intermediate signals to train SSMs.
Thus, the training dataset affects the intermediate signals through both the current data sequence and layers learned past data sequences.
In other words, even if the intermediate signals are not optimal before training, 
the training dataset can make layers generate the optimal intermediate signals through the training, which means that the training dataset has sufficient information to solve the task.
Thus, our metric is measured with parameter updates.

Algorithm~\ref{AllOne} is the computation of the K-spectral metric for one data sequence $\bm{u}^0_{0:T-1}$ inside the computation of stochastic gradient descent (SGD).
In this algorithm, we assume that the problem has the target output $y_t$ and loss function $\ell(\bm{y}^L_t,y_t)$ for each time step $t$.
Lines~\ref{code:begin_Ksp}-\ref{code:end_Ksp} are the computation of the K-spectral metric for one data sequence.
Lines~\ref{code:beginFF}-\ref{code:endFF} are the forward propagation and computation of the training loss $\mathcal{L}(\bm{\theta})=\sum_t \ell(\bm{y}^L_t,y_t)$.
In this computation, the intermediate signal $u^{l,i}_{0:T-1}$ for each SSM is stored, 
and we apply FFT to the normalized signals $\bar{u}^{l,i}_{0:T-1}$ in Line~\ref{code:FFT}.
In Line~\ref{code:eachKsp}, the K-spectral metric is computed for each signal and Line~\ref{code:AvgKsp} averages it over SSMs.
Finally, our metric is averaged over data sequences in Line~\ref{code:DataAvgKsp} of Algorithm~\ref{AllOne}, 
and thus, it is affected by the updates of weights for mini-batchs. 
This enables us to evaluate the training data 
to take into account the effect of training the learnable layers before SSMs.\looseness=-1

We recommend using the K-spectral metric after the first epoch since 
the training dataset needs to be evaluated as early as possible.
Experiments show that the K-spectral metric at the first epoch highly correlates with the performance.
If we use the K-spectral metric for another objective, e.g., using the metric for active learning,
the K-spectral metric at another epoch might be useful.
Though we do not evaluate our metric in such tasks,
we evaluate how our metric at each epoch correlates with the performance in Section~\ref{EpSec}.

\begin{algorithm}[bt]
    \caption{K-spectral metric with SGD for one epoch}
    \label{AllOne}
    \begin{algorithmic}[1]
        \Require Parameters $\bm{\theta}$, Dataset $\mathbb{D}$, learning rate $\eta$, $K$
        \State Initialization: $\mathbb{D}_e=\mathbb{D}$, $\bar{R}=0$, $\bar{\mathcal{L}}=0$
        \While{$\mathbb{D}_e\neq \varnothing$}
        \State Select a minibatch $\mathbb{B}$ from $\mathbb{D}_e$, and $\mathbb{D}_e=\mathbb{D}_e\backslash \mathbb{B}$
        \For{$(\bm{u}^0_{0:T-1},\bm{y}_{0:T-1})_n \in \mathbb{B}$}
            \State Initialization: $\mathcal{L}^n(\bm{\theta})\!=\!0$, $\bar{R}^n\!=\!0$, $N_{\mathrm{ssm}}\!=\!0$, $\bm{y}_t^0\!=\!\bm{u}_t^0$\label{code:begin_Ksp}
            \For{$t=0,\dots T-1$}
            \For{$l=1,\dots, L$}\label{code:beginFF}
            \State $\bm{u}^{l}_t=\bm{\phi}_{l-1}(\bm{y}^{l-1}_t)$
            \If{If the $l$-th layer is SSM}
            \State $\bm{y}^{l}_t,\bm{x}^{l}_{t}=\bm{\mathrm{SSMs}}(\bm{x}^{l}_{t-1},\bm{u}^{l}_t,\bm{\theta}^l)$
            \Else
            \State $\bm{y}^l_t=\bm{\phi}_l(\bm{u}^{l}_t,\bm{\theta}^l)$
            \EndIf
            \EndFor
            \State $\mathcal{L}^n(\bm{\theta})=\mathcal{L}^n(\bm{\theta})+\ell(\bm{y}^L_t,y_t)$\label{code:endFF}
            \EndFor
            \For{$l=1,\dots,L$}
            \If{If the $l$-th layer is SSM}
            \For{$i=1,\dots, d_\mathrm{in}$}
            \State $\bar{\mathcal{U}}^{l,i}_s=\mathrm{FFT}([\bar{u}^{l,i}_0,\dots,\bar{u}^{l,i}_{T-1}])$\label{code:FFT}
            \State $R(\bar{\bm{\mathcal{U}}}^{l,i}_T,K)=\sum_{k\in \mathrm{topk}(\bar{\bm{\mathcal{U}}}^{l,i}_T,K)}|\bar{\mathcal{U}}^{l,i}_k|$\label{code:eachKsp}
            \State $\bar{R}^n=\frac{1}{N_{\mathrm{ssm}}+1}(N_{\mathrm{ssm}}\bar{R}^n+R(\bar{\bm{\mathcal{U}}}^{l,i}_T,K))$\label{code:AvgKsp}
            \State $N_{\mathrm{ssm}}=N_{\mathrm{ssm}}+1$
            \EndFor
            \EndIf
            \EndFor\label{code:end_Ksp}
            \State $\bar{\mathcal{L}}=\bar{\mathcal{L}}+\frac{1}{T}\mathcal{L}^n(\bm{\theta})$
            \State $\bar{R}=\bar{R}+\bar{R}^n$
        \EndFor
        \State $\bm{\theta}=\bm{\theta}-\eta\nabla_{\bm{\theta}}\frac{1}{|\mathbb{B}|}\bar{\mathcal{L}}$
        \EndWhile
        \State $\bar{R}=\frac{1}{|\mathbb{D}|}\bar{R}$\label{code:DataAvgKsp}
        \State \textbf{Return} $\bm{\theta}$, $\bar{R}$
    \end{algorithmic}
\end{algorithm}

\section{Experiments}\label{experiments}
We conducted experiments to evaluate the effectiveness of the K-spectral metric in system identification, classification, and forecasting of time series data.
If our metric at the early epoch highly correlates with the test performance 
after training, it can evaluate the effectiveness of training data before full-training.
\looseness=-1

First, we investigate the effectiveness of the proposed metric through nonlinear identification problem using S4 in Section~\ref{SIDSec}.
We generated various input signals for identification since we need to apply the input signals and observe output signals for identification problems as explained in Section~\ref{OptInp}.
Next, we solved the classification task and forecasting tasks on public time-series data in Section~\ref{TSSec}.
We made various reduced training datasets and evaluated the correlation between the performance and the K-spectral metric.
Finally, we investigate behavior of the K-spectral metric in training and the effect of $K$ in Section~\ref{CharaSec}.

\subsection{Common setup}
We set $K$ to the number of states $d$.
\subsubsection{Models}
We used S4~\cite{S4} for system identification, classification, and forecasting problems and used
S5~\cite{S5} for classification problems.
For the system identification problem, we used three-layer deep SSMs, which are composed of one SSM layer and two input and output linear layers with SiLU activation functions.
The length of state vectors of all layers was set to four.
For classification and forecasting problems,
we used the public codes\footnote{https://github.com/state-spaces/s4}\footnote{https://github.com/lindermanlab/S5} provided by the authors of \cite{S4} and \cite{S5}.
We used S4 with Informer~\cite{Informer},
which is also included in the code, for forecasting tasks.
Hyperparameters of these experiments are the same as those in these codes.
Note that S5 was not evaluated in the forecasting tasks in the original paper~\cite{S5}, and we did not use it.
Hyperparameters of SSM layers are varied for each task, 
and they are listed in Appendix~\ref{HPModelSec}.
\subsubsection{Baseline Methods}
As a baseline, we used the number of data samples (dataset size) and validation loss after the first epoch.
The relationship between performance and the dataset size can fit a power law function~\cite{rosenfeld2020a,bahri2021explaining,mahmood2022much}.
While dataset size is known before training,
the proposed method requires training for one epoch. 
As a such baseline, we used validation loss at the first epoch as another baseline metric for evaluating training datasets. 
If the training dataset is insufficient, 
it is not known whether these baseline metrics are valid.\looseness=-1
\subsubsection{Metric for Evaluation of Metrics}
To evaluate the K-spectral metric $\bar{R}$, we used correlation coefficients between
the performance metric and $\bar{R}$.
For example, in a classification problem with six sets of [20\%, 40\%, 60\%, 80\%, 95\%, 99\%] reduced datasets, 
we compute
\begin{align}\textstyle
    \rho = \frac{\sum_i{(Acc_i - \frac{1}{N_{\mathrm{d}}}\sum_i{Acc_i})(\bar{R}_i - \frac{1}{N_{\mathrm{d}}}\sum_i{\bar{R}_i})}}{\sqrt{\sum_i{(Acc_i - \frac{1}{N_{\mathrm{d}}}\sum_i{Acc_i})^2}\sum_i{(\bar{R}_i -\frac{1}{N_{\mathrm{d}}}\sum_i{\bar{R}_i})^2}}},\label{coeffeq}
\end{align}
where $Acc_i$ and $\bar{R}_i$ are test accuracy and the proposed metric for the $i$-th dataset, respectively.
$N_\mathrm{d}$ is the number of datasets, i.e., $N_{\mathrm{d}}\!=\!6$ in this example.
System identification problems and forecasting tasks use mean squared error (MSE) instead of accuracy.\looseness=-1
\subsection{System Identification}\label{SIDSec}
The major difference between system identification and machine learning is that system identification involves creating a training dataset, 
where we apply input signals to a physical system and observe its output signals. 
The optimal input design is not obvious when the target is an unknown nonlinear system. 
In this experiment, we evaluate the metrics of the training dataset when we train deep SSMs for modeling two ground-truth systems that have linear dynamics and static non-linearity.
\subsubsection{Set up}
We assume that the true systems are a Wiener model~\cite{biagiola2009wiener,wigren1993recursive} and Hammerstein model~\cite{biagiola2009wiener,falugi2005approximation} with the additive Gaussian noise.
The Wiener-model has the static nonlinear component after the linear dynamical system. 
Following \cite{biagiola2009wiener,wigren1993recursive}, we used the mathematical model for the control valve for fluid flow:
\begin{align}\textstyle
    v(t)&\textstyle=\frac{0.1044q^{-1}+0.0883q^{-2}}{1-1.4138q^{-1}+0.6065q^{-2}}u(t),\nonumber\\
    \textstyle y(t)&\textstyle=\frac{v(t)}{\sqrt{0.10+0.90v^2(t)}}\label{Wmodel}
\end{align}
where $u(t)$, $v(t)$, and $y(t)$ are the signal applied to the stem, the stem position, and the resulting flow, respectively.
On the other hand, the Hammerstein-model is also composed of the static nonlinear component and linear system  but nonlinearity is before the linear system. 
Following~\cite{biagiola2009wiener,falugi2005approximation},  
we used the polynomial nonlinearity with FIR:\looseness=-1
\begin{align}\textstyle
    v(t)&\textstyle=\sum_{i=1}^{3}p_i u^i (t)\nonumber\\
    y(t)&\textstyle=\sum_{i=1}^{8}\theta_i v(t-i)\label{Hmodel}
\end{align}
where $\bm{p}=[1, 3, 2]^{\mathsf{T}}$ and $\bm{\theta}=[1,2,0.3,4,1,1,1,0.5]^{\mathsf{T}}$.

Training datasets are generated by applying input signals $u(t)$ to the above models and observing $y(t)$.
We observed each input and output signal $\{(u_t,y_t)\}_{t=1}^{10000}$
and used $\{(u_t,y_t)\}_{t=1}^{8000}$ as a training dataset and $\{(u_t,y_t)\}_{t=8001}^{10000}$ as a validation dataset.
The number of time steps $T$ for the test set is also set to 10,000.
To prepare training datasets with various bias,
input signals are generated to have different frequencies as follows:
{\setlength{\leftmargini}{20pt}
\begin{description}\setlength{\topsep}{3pt}
    \item[Training datasets] 
We made 5,000 training datasets for computation of the correlation coefficients.
We generated 5,000 input signals and observed the output signal for each input signal. 
The $i$-th input signal has $i$ frequency components:
\begin{align}\textstyle
    u^0(t)=c\sum_{j=0}^{i-1} \mathrm{sin}(\frac{2\pi\omega_j t}{T}+\frac{4\pi\psi_j}{T} )\label{SIDInput}
\end{align}
where $\omega_j$ and $\psi_j$ are sampled from $U(0,T/2)$.
$c\in \mathbb{R}$ is set to satisfy $||\bm{u}^0_{0:T-1}||_2=100$.
$y_t$ is obtained by applying $u^0(t)$ to the ground-truth systems (Eqs.~(\ref{Wmodel}) and (\ref{Hmodel})).
\item[Test datasets] 
Since we do not know input signals for the running plants and controlled fluid flow in advance,
test input signals should be different from training input signals for nonlinear system identification problems.
We prepared two input signals as the test dataset.
Test input~I is the same as the input signals (\req{SIDInput}) when $i=5000$.
Test input II is a signal that takes a constant value sampled from a uniform distribution $U(-1,1)$ for each interval.
We set the length of the interval to 20 and $T$ to 10,000.
Finally, $u^0(t)$ is normalized to satisfy $||\bm{u}^0_{0:T-1}||_2\!=\!100$.
We observed the true output signals for each input signal. \looseness=-1
\end{description}
}
We evaluate MSE between the true outputs and the outputs of deep SSMs after the training of each training datasets
and compute $\rho$ with 5000 training datasets ($N_d=5000$) by \req{coeffeq}.
\begin{table}
    \caption{Correlation coefficients between MSE of test datasets and each metric on identification problems using S4.}
    \label{Identification}
    \centering
    \scalebox{0.89}{
    \begin{tabular}[tb]{ccccccccc}\toprule
        &\multicolumn{2}{c}{Wiener Model}&\multicolumn{2}{c}{Hammerstein Model}\\
        &Test input I &Test input II&Test input I&Test input II\\\midrule
        Dataset size&N/A &N/A&N/A &N/A\\
        Valid. loss&-$0.28\pm0.12$&-$0.15\pm 0.04$&-$0.14\pm0.4$&$0.0\pm 0.3$\\
        $\bar{R}$&$\bm{0.53\pm 0.09}$&$\bm{0.70\pm 0.04}$&$\bm{0.61\pm0.04}$&$\bm{0.6\pm0.2}$
        \\\bottomrule
    \end{tabular}
    }
\end{table}
\subsubsection{Results}
\rtab{Identification} lists the correlation coefficients between the
K-spectral metric and MSE for the test dataset and between baselines and MSE.
The K-spectral metric achieves the highest absolute values of correlation coefficients among the metrics.
Validation loss does not correlate with the test loss because a 
validation dataset is a subset of a training dataset: 
if a training dataset is biased and does not have sufficient information to identify nonlinear systems, 
the behavior of validation loss is different from the behavior of test loss. 
Since dataset sizes $T$ are constant across training datasets, correlation coefficients cannot be computed.
This is a drawback of using dataset size to evaluate the effectiveness of training datasets.
In contrast, the K-spectral metric can evaluate the effectiveness of training datasets 
even when the training dataset sizes are constant and the data collection is biased.
\begin{table*}[tb]
    \centering
    \caption{Correlation coefficients between test accuracy and each metric on classification problems. 
    We compute the coefficients on only (i) randomly reduced training datasets, only (ii) training datasets lacking some classes,
    and joint sets of datasets (i) + (ii). }
    \label{Classification}
    \begin{tabular}[tb]{ccccccccc}\toprule
        &Models&Metrics&CIFAR10&ListOps&Speech Commands\\\midrule
        (i) &S5 &Dataset size&$0.901\pm0.006$&$0.785\pm0.009$&$0.69\pm0.01$\\
        &&Valid. Acc.&$0.88\pm0.05$&$0.86\pm0.04$&$0.79\pm0.04$\\
        &&$\bar{R}$&$\bm{0.985\pm0.004}$&-$\bm{0.87\pm0.08}$&$\bm{0.91\pm0.01}$\\\cmidrule(l){2-6}
        &S4 &Dataset size&$0.901\pm0.005$&$\bm{0.849\pm0.005}$&$0.79\pm0.01$\\
        &&Valid. Acc.&$\bm{0.97\pm0.01}$&$0.78\pm0.03$&$\bm{0.88\pm0.02}$\\
        &&$\bar{R}$&$0.88\pm0.03$&-$0.80\pm0.03$&-$0.4\pm0.2$
        \\\midrule
        (ii)&S5 &Dataset size&$\bm{0.9991\pm0.0001}$&$\bm{0.998\pm0.002}$&$\bm{0.999988\pm7\times10^{-6}}$\\
        &&Valid. Acc.&-$0.93\pm0.01$&-$0.91\pm0.02$&$0.1\pm0.7$\\
        &&$\bar{R}$&$0.982\pm0.01$&-$0.940\pm0.05$&$0.73\pm0.07$\\\cmidrule(l){2-6}
        &S4 &Dataset size&$\bm{0.9992\pm0.0002}$&$\bm{0.999\pm0.001}$&$\bm{0.99992\pm3\times10^{-5}}$\\
        &&Valid. Acc.&-$0.93\pm0.01$&-$0.6\pm0.4$&-$0.8\pm0.1$\\
        &&$\bar{R}$&$0.993\pm0.002$&-$0.91\pm0.03$&-$0.89\pm0.01$
        \\\midrule
        (i) + (ii)&S5 &Dataset size&$0.70\pm0.02$&$0.525\pm0.008$&$0.338\pm0.006$\\
        &&Valid. Acc.&-$0.36\pm0.06$&-$0.17\pm0.1$&-$0.20\pm0.05$\\
        &&$\bar{R}$&$\bm{0.77\pm0.01}$&-$\bm{0.6\pm0.1}$&$\bm{0.44\pm0.01}$\\\cmidrule(l){2-6}
        &S4 &Dataset size&$0.64\pm0.02$&$0.48\pm0.01$&$0.37\pm0.01$\\
        &&Valid. Acc.&-$0.38\pm0.01$&-$0.3\pm0.1$&-$0.14\pm0.03$\\
        &&$\bar{R}$&$\bm{0.84\pm0.02}$&-$\bm{0.809\pm0.007}$&-$\bm{0.49\pm0.06}$
        \\\bottomrule
    \end{tabular}
\end{table*}
\subsection{Classification and Forecasting}\label{TSSec}
\subsubsection{Setup}
We used CIFAR10~\cite{cifar}, ListOps~\cite{ListOps}, and Speech Commands (SC)~\cite{SC} for classification problems of time-series data following \cite{S4,S5}. 
To evaluate metrics on datasets with various bias, we prepared (i) randomly reduced datasets and (ii) datasets lacking certain class data. 
For (i), we used [20\%, 40\%, 60\%, 80\%, 95\%, 99\%] of the training datasets.
Validation datasets are subsets of these reduced training datasets.
For (ii), we first apply validation split following the previous works~\cite{S4,S5}
and randomly removed $[2, 4, \dots, 10]$ target labels and their data for CIFAR10 and ListOps,
and $[7,14,\dots,35]$ target labels for SC.
In all cases, the full test datasets were used for the performance metrics. 
Thus, the last layer of the model was designed to produce outputs for all classes even when certain classes were missing from the training dataset.

For forecasting tasks, we used Weather, ECL, $\mathrm{ETTh}_1$, and $\mathrm{ETTm}_1$ \cite{Ett,Ecl} following~\cite{S4}.
In the forecasting tasks, we only used randomly reduced datasets because it is difficult to remove data on the basis of a certain target labels from regression tasks. 
Additionally, since the training dataset for the forecasting task specifies a specific interval of a continuous dataset,
it was difficult to reduce training datasets in the same way as classification problems.
Several intervals were randomly removed from each training dataset to make them as equally sized as possible.
As a result, sizes for the datasets were as following:
[3000, 6500, 10000, 13500, 17050, 20550, 24050] for Weather,
[2150, 4750, 7400, 10000, 12650, 15300, 17900] for ECL, 
[950, 1650, 2400, 3800, 4550, 5250, 6000] for $\mathrm{ETTh}_1$,
[5250, 8150, 11000, 16800, 19650, 22550, 25400] for $\mathrm{ETTm}_1$.
The original full test datasets were used for the performance metrics.\looseness=-1 
\subsubsection{Correlation between K-spectral Metric and Performance}\label{MainEx}
\rtab{Classification} lists the correlation coefficients between test accuracy and each metric for classification problems,
and \rtab{Forecasting} lists the correlation coefficients between test MSE and each metric for forecasting problems.\looseness=-1

In \rtab{Classification}, we compute correlation coefficients $\rho$ on three sets of training datasets:
(i) randomly reduced datasets, (ii) datasets lacking certain classes, and the joint set of (i) and (ii).
Regarding dataset size, the correlation coefficients are large when using only (i) and only (ii).
Especially, the relationship between dataset size and the performance of (ii) is almost linear.
This is because test accuracy linearly decreases when training datasets lack classes one by one.
However, for the joint set of training datasets (i)+(ii), dataset size loses the linear correlation.
Additionally, dataset size does not necessarily correlate highly with the performance for only (i)
because the relation is a power law rather than a linear correlation~\cite{rosenfeld2020a}.
Validation accuracy moderately correlates with the performance in the case of (i).
However, it is not very useful metric in the case of (ii) because
validation datasets are affected by the lack of information for the task in training datasets. As a result, it does not correlate with the performance to evaluate the joint set.

On the other hand, our metric has large absolute values of correlation coefficients ($>$ 0.7) in most cases
when using only (i) or (ii). 
Though the absolute values of coefficients are not always larger than those of dataset size and validation accuracy,
this result indicates that the spectra of intermediate signals are correlated with the performance.
Furthermore, for the joint set of training datasets (i)+(ii), 
our metric achieves the highest absolute values of correlation coefficients.
Since practical data analyses can suffer from mixed quality such as 
training datasets lack information to solve the task uniformly and biasedly, 
this result implies our metric helps data collection of new tasks of time-series data analysis applications.

The K-spectral metric can have negative correlation coefficients.
This implies that SSMs should have non-uniform sensitivity in the frequency domain for good performance 
because the K-spectral metric becomes large when target SSMs have sensitivity uniformly in frequency domain as discussed in Section~\ref{RKOptIntSec}.
Even so, since the absolute values are larger than those of dataset size, 
we can use the K-spectral metric as the performance metric after observing at least two datasets.

\rtab{Forecasting} lists the correlation coefficients of metrics in forecasting problems.
The dataset size and the performance are negatively correlated
since a smaller MSE indicates a better performance. 
On the other hand, the proposed metric and validation loss are positively correlated.
The absolute values of their correlation coefficients are larger than those of the dataset size.
This indicates that the validation loss and proposed metric are linearly correlated with the performance.
This result also supports that our metric can evaluate the effectiveness of training data and that 
its effectiveness is comparable to the validation loss when training datasets can uniformly lack the information to solve the task.
Since our metric is computed on training datasets at the first epoch, we can estimate the test performance by using our metric without full-training.
This can accelerate the practical use of deep SSMs on time-series data.\looseness=-1
\begin{table}
    \caption{Correlation coefficients between test accuracy and each metric on forecasting problems using S4. 
    To calculate the coefficients, we used randomly sampled training datasets.}
    \label{Forecasting}
    \centering
    \scalebox{0.84}{
    \begin{tabular}[tb]{ccccccccc}\toprule
        Datasets&Weather&ECL&$\mathrm{ETTh}_1$&$\mathrm{ETTm}_1$\\\midrule
        Dataset size&-$0.632\pm0.005$&-$0.71\pm0.02$&-$0.782\pm0.008$&-$0.85\pm0.01$\\
        Valid. loss&$\bm{0.95\pm0.06}$&$\bm{0.9\pm0.1}$&\bm{$0.993\pm0.02}$&$\bm{0.943\pm0.02}$\\
        $\bar{R}$&$0.94\pm0.03$&$\bm{0.9\pm0.1}$&$0.8\pm0.1$&$0.93\pm0.03$
        \\\bottomrule
    \end{tabular}
    }
\end{table}
\subsection{Characteristics of K-spectral Metric}\label{CharaSec}
In this section, we investigate the characteristics of the K-spectral metric in detailed.
We used S5 with (i) randomly reduced training datasets
on classification problems for computation of $\rho$. 
\subsubsection{How does K-spectral Metric Change in Training?}\label{EpSec}
\begin{figure}[tb]
    \centering
    \subfloat[][CIFAR10]{\includegraphics[width=0.33\linewidth]{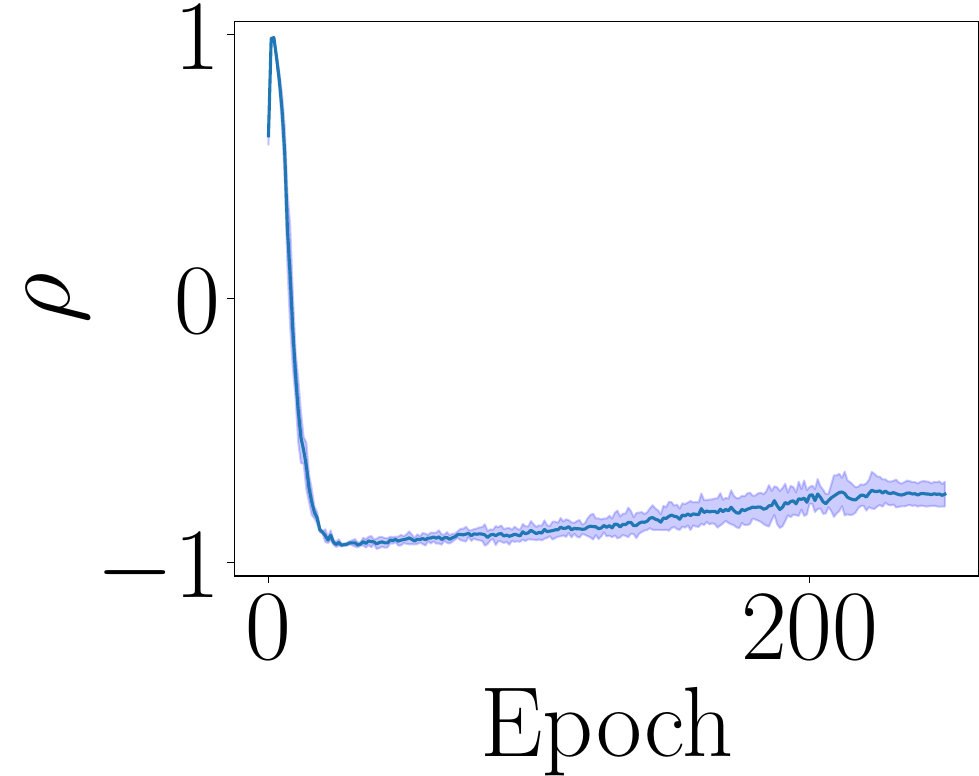}\label{C10Ep}}
    \subfloat[][ListOps]{\includegraphics[width=0.33\linewidth]{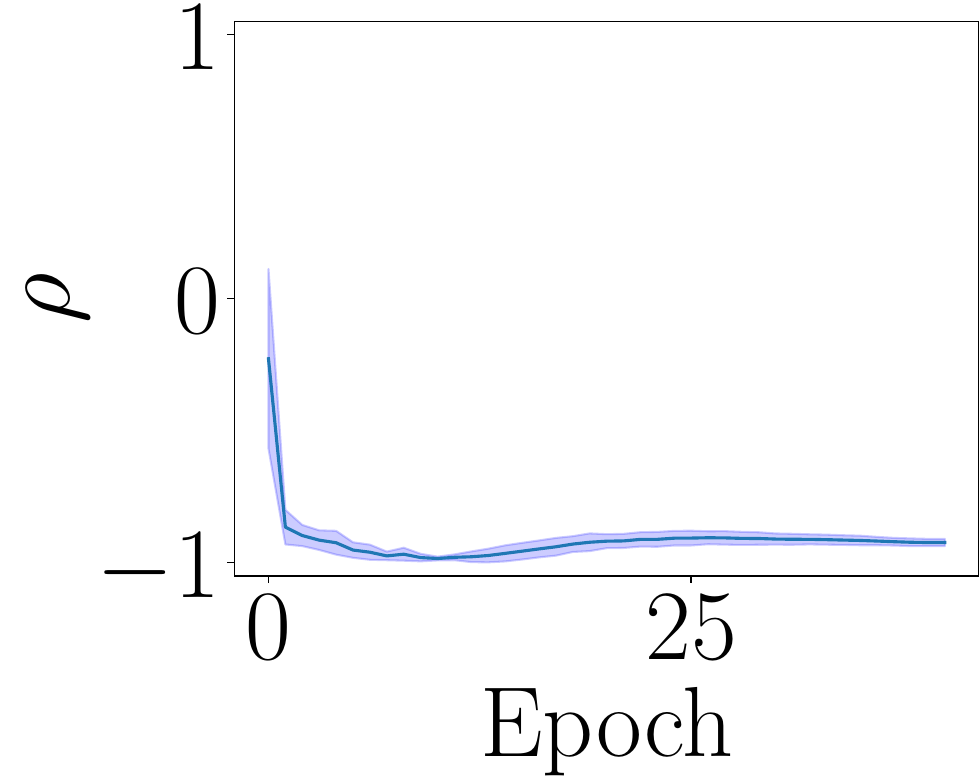}\label{ListOpsEp}}
    \subfloat[][SC]{\includegraphics[width=0.33\linewidth]{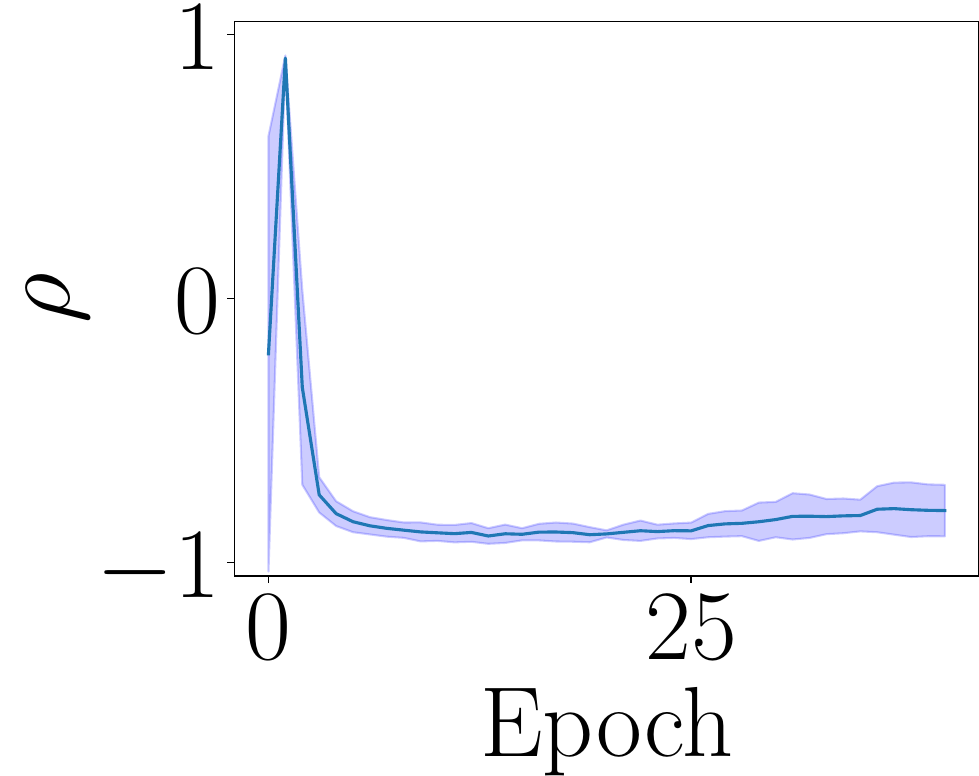}\label{SCEp}}
    \vspace{-5pt}\caption{$\rho$ of $\bar{R}$ against Epochs in training S5 on (i).}
    \label{EpEx}
    \centering
    \subfloat[][CIFAR10]{\includegraphics[width=0.33\linewidth]{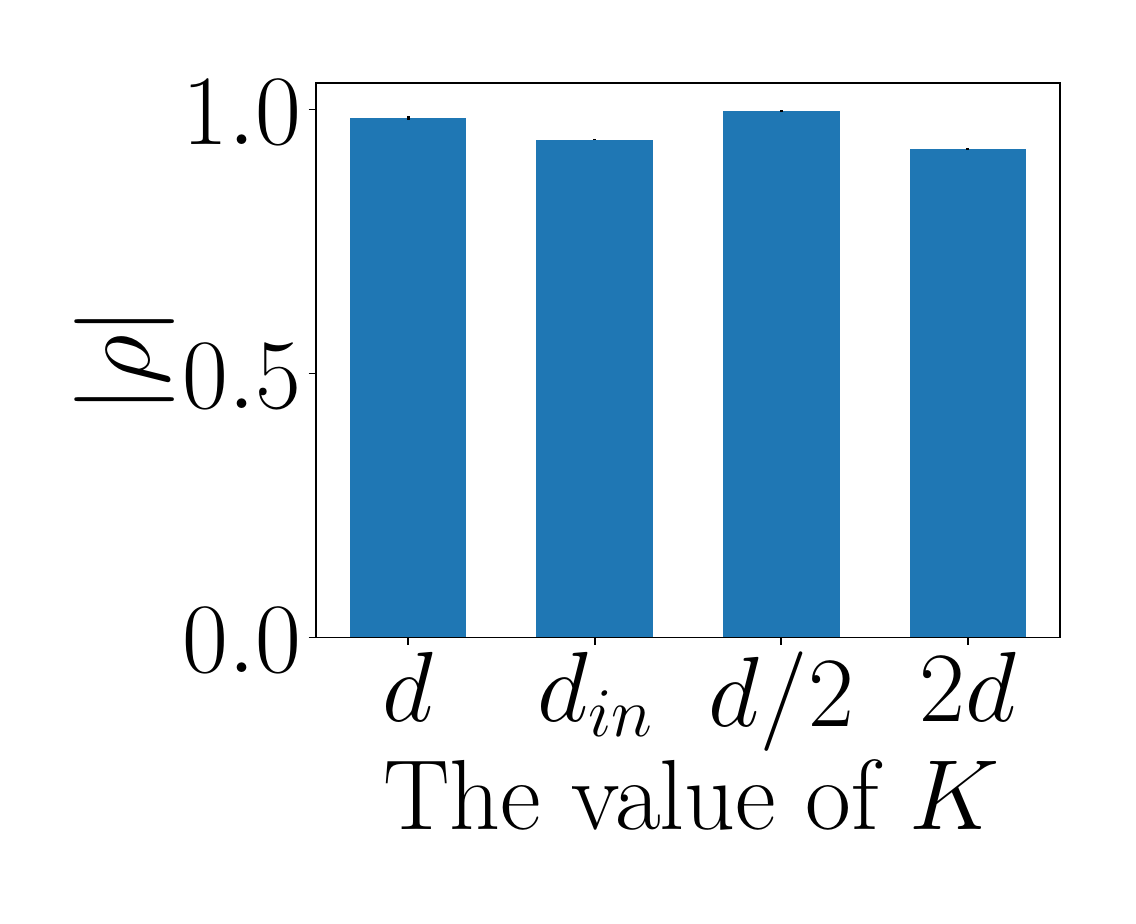}\label{C10TopK}}
    \subfloat[][ListOps]{\includegraphics[width=0.33\linewidth]{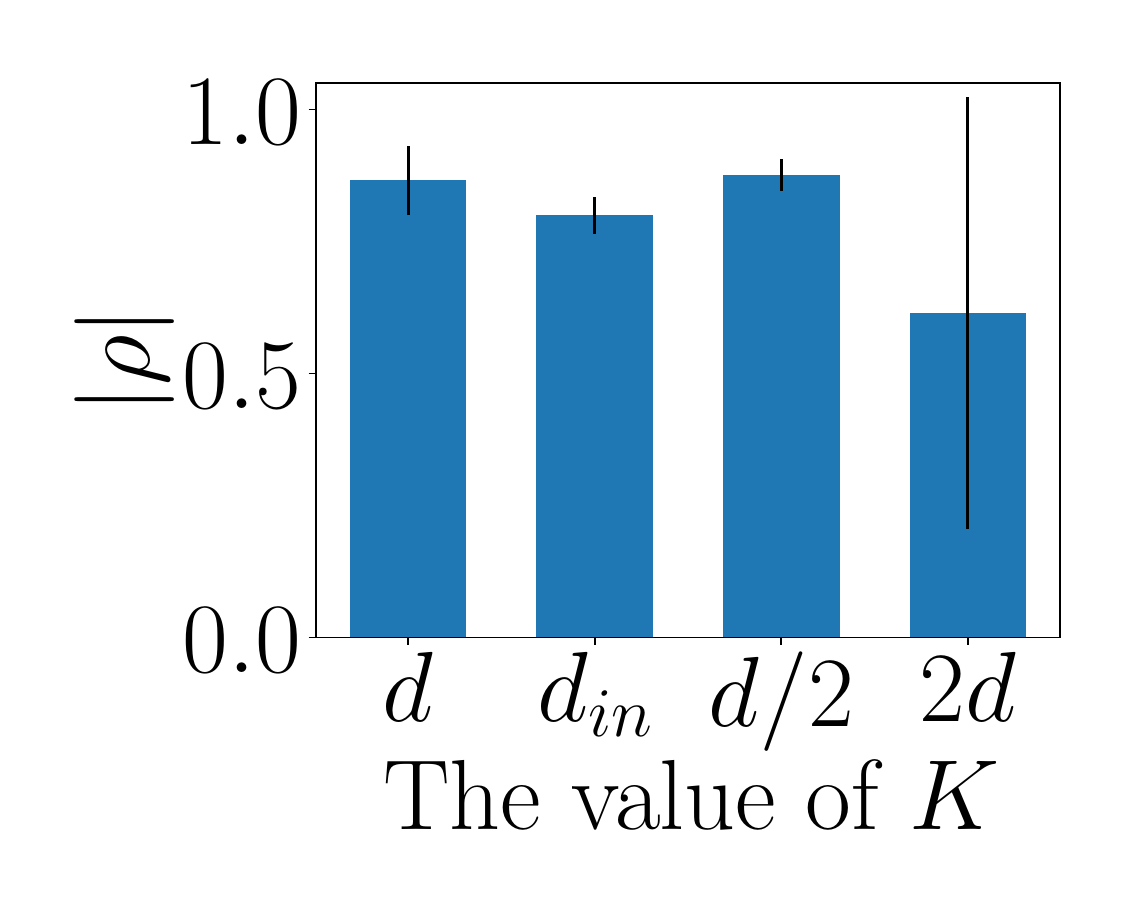}\label{ListOpsTopK}}
    \subfloat[][SC]{\includegraphics[width=0.33\linewidth]{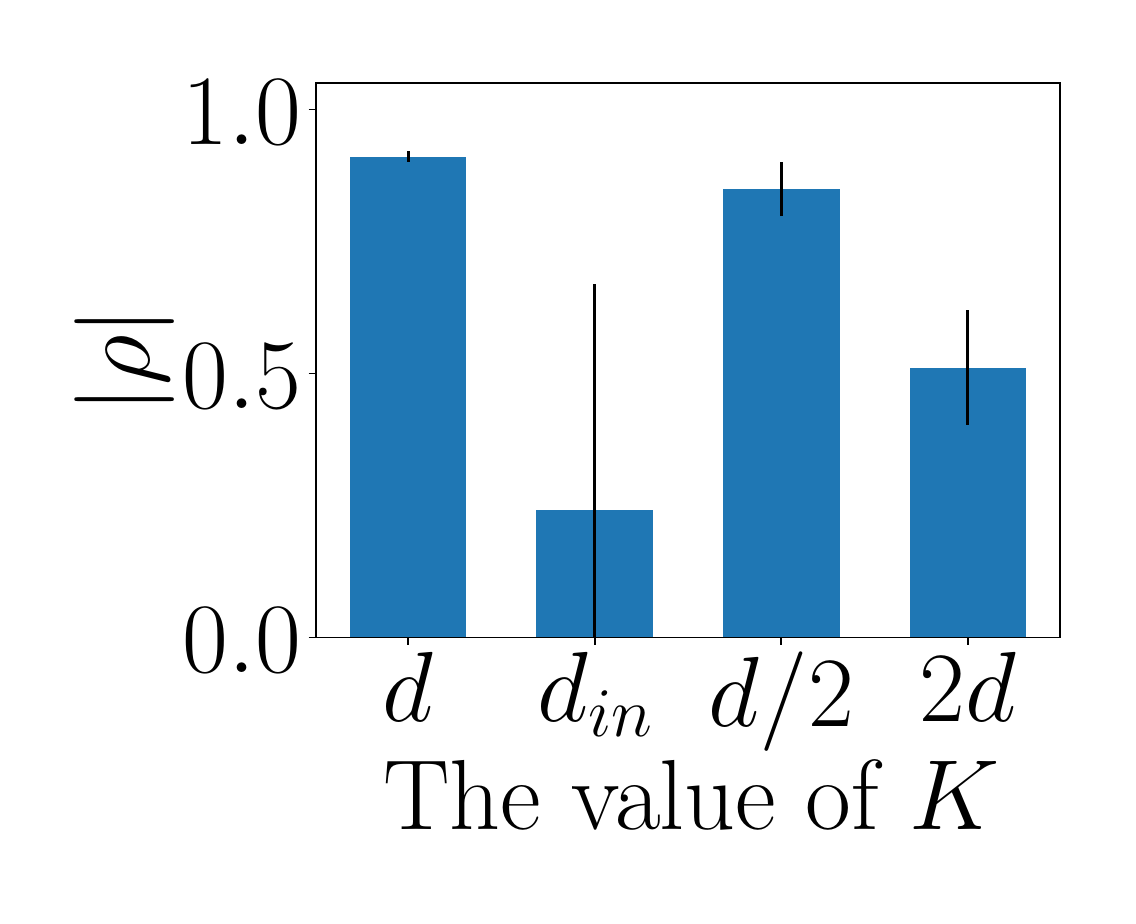}\label{SCTopK}}
    \caption{$|\rho|$ of $\bar{R}$ against $K$ in training S5 on (i).}
    \label{TopKEx}\vspace{-10pt}
\end{figure}

As mentioned in Section~\ref{Impl}, $u^l_t$ is affected by training of the $l^\prime\!<\!l$-th learnable layers.
This also affects the performance, and thus, our metric should consider this effect.
Thus, we evaluate our metric against epoch. \looseness=-1
\rfig{EpEx} plots the correlation coefficients between $\bar{R}$ and the test accuracy in classification problems using S5.
For the 0-th epoch, we evaluate $\bar{R}$ of deep SSMs with random weights before training.
Early in the training (the 1st-5th epochs), correlation coefficients tend to have large values.
Especially, on CIFAR10 and SC, the correlation coefficients achieve about one.
This indicates that the flat spectrum early in the training is necessary
to train S5. 
Intriguingly, the correlation coefficients are about -1.0 in the middle of training.
This indicates that intermediate signals should have non-uniform spectra like \rfig{Sig4}.
This implies that there is a specific important frequency area in the frequency response of SSMs to solve the tasks.

Since the absolute values are almost one, the performance and our metric have a linear relationship.
This result indicates that 
\textit{even if SSMs are inside DNNs, we can use the spectrum 
to evaluate the datasets like the concept of optimal input design and PE}.
Unlike linear system identification, 
datasets are difficult to optimize to maximize or minimize the K-spectral metric.
Even so, we might use this metric for datasets comparison, the evaluation of data augmentations, or active learning.
The evaluation of training datasets at just one epoch instead of full-training enables
efficient development of the time-series data analysis applications with deep SSMs.
\subsubsection{Effect of $K$}\label{KEx}
In the above experiment, we set $K$ to the number of the states $d$
because the identification problem of a SISO SSM with $d$ states requires the PE of order at least $d$.
This section gives the results of other settings $K\!=\!d_{\mathrm{in}}$, $d/2$, and $2d$.
In most cases, $d_{\mathrm{in}}$ is larger than $d$ (Appendix~\ref{HPModelSec}).
\rfig{TopKEx} plots the absolute values of correlation coefficients $|\rho|$ against $K$.
Since $K\!=\!d$ has the largest or the second largest value of $|\rho|$,
our setup policy of $K$ works well.
The setting of $K\!=\!d/2$ also has large $|\rho|$. This implies that the number of important frequencies might be less than $d$ for these tasks. 
The setting of $K\!=\!2d$ is not good choice although it is larger than $d$. 
This is because PE of higher order is not necessarily better once the PE condition is satisfied.
In addition, K-spectral metric becomes larger when signals have the flat spectrum, 
but the optimal spectrum can be non-uniform as discussed in Section~\ref{EpSec}.
Note that $\rho$ has a negative value in some cases, and we used $|\rho|$ to improve visibility.
\section{Conclusion}
In this paper, we investigate the relationship between the performance of deep neural networks with state space models (deep SSMs) and frequency components of intermediate signals.
We proposed the K-spectral metric on the intermediate signals 
to evaluate a training dataset.
Experiments revealed that, although deep SSMs are nonlinear systems,
the K-spectral metric is highly correlated with the performance.
Since our metric does not consider the desired SSMs after training,
it has high value for the flat spectrum whereas the optimal input design considers sensitivity of linear systems.
In future work, we will investigate an evaluation metric considering the suitable SSMs.

\bibliography{CameraBib.bib}
\bibliographystyle{ACM-Reference-Format}
\appendix
\section{Proof of Theorem~\ref{thm}}\label{ProofApp}
We proved Theorem~\ref{thm} as follows:
\begin{proof}
    From Parseval's theorem, we have
    $\sum_{t}|u_t|^2\!=\!\frac{1}{T}\!\sum_{s}|\mathcal{U}_s|^2$.
    Thus, we have $\sum_{s}|\bar{\mathcal{U}}_s|^2\!=\!T$ for the normalized signal $\bar{u}_t$.
    We consider the maximization problem:
    \begin{align}\textstyle
        \bar{\bm{\mathcal{U}}}^*_T=\mathrm{arg}\!\max_{\bar{\bm{\mathcal{U}}}_T}\sum_{k\in \mathrm{topk}(\bar{\bm{\mathcal{U}}}_T,K)}|\bar{\mathcal{U}}_k|\\
       \textstyle\mathrm{subject~to}~~\sum_{s=0}^{T-1}|\bar{\mathcal{U}}_s|^2=T.
    \end{align}
    From the constraint, it is obvious that
    the solution of $\bar{\bm{\mathcal{U}}}^*_T$ satisfies $|\bar{\mathcal{U}}^*_i|\!=\!0$ for $i \notin \mathrm{topk}(\bar{\bm{\mathcal{U}}}_T,K)$.
    Let $\bm{a}=[a_1,\dots,a_K]^\mathsf{T}$ be a vector composed of the top $K$ of $|\bar{\mathcal{U}}_s|$.
    This problem can be written the problem:
    $\bm{a}^*=\mathrm{arg}\!\max_{\bm{a}}\sum_{k=1}^{K}a_k$
     subject to $\sum_{k=1}^{K}a_k^2=T$ and $a_k\geq 0$. 
     It is easily solved by the method of Lagrange multiplier, and
     the solution is $a^*_i=\sqrt{T/K}$ for $\forall i$, which completes the proof.
    \end{proof}

\section{Derivation of Objective Function for Optimal Input Design}\label{FIMExApp}
We explain the objective function of the optimal input design following~\cite{rojas2007robust}.
We consider the system identification of the following system:
\begin{align}\label{discTrans}
    y_t=G_1(q,\bm{\theta})u_t+G_2(q,\bm{\gamma})e_t,
\end{align}
where $G_1(q,\bm{\theta})$ and $G_2(q,\bm{\gamma})$ is rational transfer functions
for a discrete system and noise, respectively. $q$ is shift operator as $qu_t=u_{t+1}$ and $q^{-1}u_t=u_{t-1}$.
$\bm{\theta}$ and $\bm{\gamma}$ are parameters.
$e_t$ is zero mean Gaussian white noise of variance $\sigma$.
For example, FIR of \req{fir} is written as:
\begin{align}
    G_1(q,\bm{\theta})=\sum_{i=1}^{d}\theta_i q^{-i},\\
    G_2(q,\bm{\gamma})=1,
\end{align}
and $G_2(q,\bm{\gamma})=1$ is also assumed in \req{FIM}.
By applying FFT to the both sides of \req{discTrans}, we have
\begin{align}\label{FFTTrans}
    y({e^{j\omega T}})=G_1(e^{j\omega T},\bm{\theta})u({e^{j\omega T}})+G_2(e^{j\omega T},\bm{\gamma})e(e^{j\omega T}).
\end{align}
Let $\bm{y}_{1:T}$ be observed outputs $\bm{y}_{1:T}=[y_1,\dots,y_T]$, we have the following log likelihood:
\begin{align}\label{LL}
    \log p(\bm{y}_{1:T}|\bm{\beta},\bm{u}_{1:T})=-\frac{T}{2}\log 2\pi-\frac{T}{2}\log \sigma-\frac{1}{2\sigma}\sum_{t=1}^{T}\varepsilon_t^2,
\end{align}
where $\bm{\beta}=[\bm{\theta}^{\mathsf{T}},\bm{\gamma}^{\mathsf{T}},\sigma]^{\mathsf{T}}$.
Note that $p(\bm{y}_{1:T}|\bm{\beta})$ is a Gaussian distribution because \req{discTrans} is a linear dynamical system and stochastic element is only $e_t$.
$\varepsilon_t$ is written as follows:
\begin{align}\label{eps}
\varepsilon_t = G_2 (q,\bm{\gamma})^{-1}\left[ y_t - G_1(q,\bm{\theta} ) u_t \right].   
\end{align}
Fisher information matrix for \req{LL} is given by 
\begin{align}\label{rawFIM}
    \bm{M}\!=\!\mathbb{E}_{\bm{y}_{1:T}|\bm{\beta}}\!\!\left[\!\left(\frac{\partial\log p(\bm{y}_{1:T}|\bm{\beta})}{\partial \bm{\beta}}\right)\!\!\left(\frac{\partial\log p(\bm{y}_{1:T}|\bm{\beta})}{\partial \bm{\beta}}\!\right)^{\mathsf{T}}\right].
\end{align}
This matrix bounds the variance-covariance matrix of unbiased estimators $\bm{\beta}$ from the Cram\'er Rao bound.
From Eqs.~(\ref{LL}) and (\ref{eps}), we have
\begin{align}
    \frac{\partial \varepsilon_t}{\partial \bm{\beta}}=-G_2 (q,\bm{\gamma})^{-1}\left[ \frac{\partial G_2(q,\bm{\gamma})}{\partial \bm{\beta}}\varepsilon_t+\frac{\partial G_1(q,\bm{\theta} )}{\partial \bm{\beta}} u_t \right].
\end{align}
We assume that $e_t$ and $u_t$ are uncorrelated. Then, \req{rawFIM} is written as
\begin{align}
 \bm{M}=\left[\begin{smallmatrix}
    \bm{M}_1&\bm{O}\\
    \bm{O}&\bm{M}_2
 \end{smallmatrix}\right],  
\end{align}
where $\bm{M}_1$ is the part of the information matrix depending on the input signals.
$\bm{M}_2$ is independent of the input signals, and thus, it cannot modified by the input design.
$\bm{M}_1$ is written as 
\begin{align}
    \bm{M}_1=\frac{1}{\sigma}\sum_{t=1}^{T}\left(\frac{\partial \varepsilon_t}{\partial \bm{\theta}}\right)\left(\frac{\partial \varepsilon_t}{\partial \bm{\theta}}\right)^{\mathsf{T}}.
\end{align}
For the sufficiently large $T\rightarrow \infty$,
we have the following equality from the Parseval's theorem:
\begin{align}
    \bar{\bm{M}}=\lim_{T\rightarrow\infty}\frac{1}{T}\bm{M}_1\sigma=\frac{1}{\pi}\int_{0}^{\pi}\tilde{\bm{M}}(\bm{\beta},\omega)\Phi_u(\omega)d\omega,
\end{align}
where 
\begin{align}\label{rawtilM}
    &\tilde{\bm{M}}(\bm{\beta},\omega)\\
    &=\mathrm{Re}\left\{\frac{\partial G_1(e^{j\omega},\bm{\theta} )}{\partial \bm{\theta}}|G_2(e^{j\omega},\bm{\gamma} )|^{-2} \left[\frac{\partial G_1(e^{j\omega},\bm{\theta} )}{\partial \bm{\theta}}\right]^H \right\}.
\end{align}
Since we assume $G_2(q,\bm{\gamma})=1$ in the paper,
\req{rawtilM} becomes 
\begin{align}
    \tilde{\bm{M}}(\bm{\theta},\omega)=\mathrm{Re}\left\{\frac{\partial G_1(e^{j\omega},\bm{\theta} )}{\partial \bm{\theta}}\left[\frac{\partial G_1(e^{j\omega},\bm{\theta} )}{\partial \bm{\theta}}\right]^H \right\}.
\end{align}
Since it is difficult to use the matrix for the optimization problem,
the scalar function of $\bar{\bm{M}}$ is used for the optimal input design.

We assume that there exists the true deep SSMs $F(\bm{X}_{t},\bm{u}_t,\bm{\theta}^*)$ for a task,
 and dataset is generated by the deterministic state transition with stochastic components $\bm{e}_t$:
\begin{align}
    \bm{y}^{L}_t=F(\bm{X}_{t-1}+\bm{e}_{t},\bm{u}^0_t, \bm{\theta}^*).
\end{align}
where $\bm{X}$ is concatenated state vectors of intermediate layers.
Additionally, we assume that transition functions of the state vectors are SSM layers with additive Gaussian noise $e_t\sim \mathcal{N}(0,\sigma)$:
$ y^{l}_t=\bm{c}^{l\mathsf{T}*}\bm{x}^{l}_t+D^{l*}u^{l}_t+e_t$.
Note that these assumptions do not mean that 
the output of the deep SSMs $\bm{y}^{L}_t$ follows normal distribution because
deep SSMs contain nonlinearity before and after SSMs: $\bm{y}^{L}_t$ can follow complicated distributions even if the intermediate state vectors follow normal distributions. 
Under these conditions, 
$\log p(\bm{y}^l_{1:T}|\bm{\beta}^l,\bm{u}^l_{1:T})$, which is the likelihood  of the output signals of the $l$-th SSM inside deep SSMs,
 is given by \req{LL} where $\bm{\beta}^l=[\mathrm{vec}(\bm{A}^l)^\mathsf{T},\bm{b}^{l\mathsf{T}},\bm{c}^{l\mathsf{T}},D^{l},\sigma]^\mathsf{T}$.
Thus, if training of deep SSMs can be considered as unbiased estimator of parameters of the $l$-th SSM $(\bm{A}^l,\bm{b}^l,\bm{c}^l,D^l)$,
\req{rawFIM} can be used to bound the variance of the parameters of the $l$-th SSM.
Therefore, we can apply the optimal input design to identify SSMs inside DNNs under these assumptions.
\section{Hyperparameters of S4 and S5}
\label{HPModelSec}
Hyperparameters of S4 and S5 are listed in \rtab{HPModel}.
The settings of classification and forecasting are based on the public code of S4 and S5.
\begin{table}[tb]
    \centering
    \caption{Main hyperparameters of SSMs.}
    \label{HPModel}
    \begin{tabular}[tb]{ccccccccc}\toprule
        &Models&$d$&$d_{in}$&\# of layers\\\midrule
        System identification &S4 &4&4&1\\\midrule
        &S4 &64&512&6
        \\\midrule
        ListOps&S5&16&128&8 \\
        &S4 &4&256&6
        \\\midrule
        SC&S5 &128&96&6\\
        &S4 &64&128&6\\\midrule
        Forecasting tasks&S4 &64&128&2
        \\\bottomrule
    \end{tabular}
\end{table}
\begin{figure}[tb]
    \centering
    \subfloat[][CIFAR10]{\includegraphics[width=0.33\linewidth]{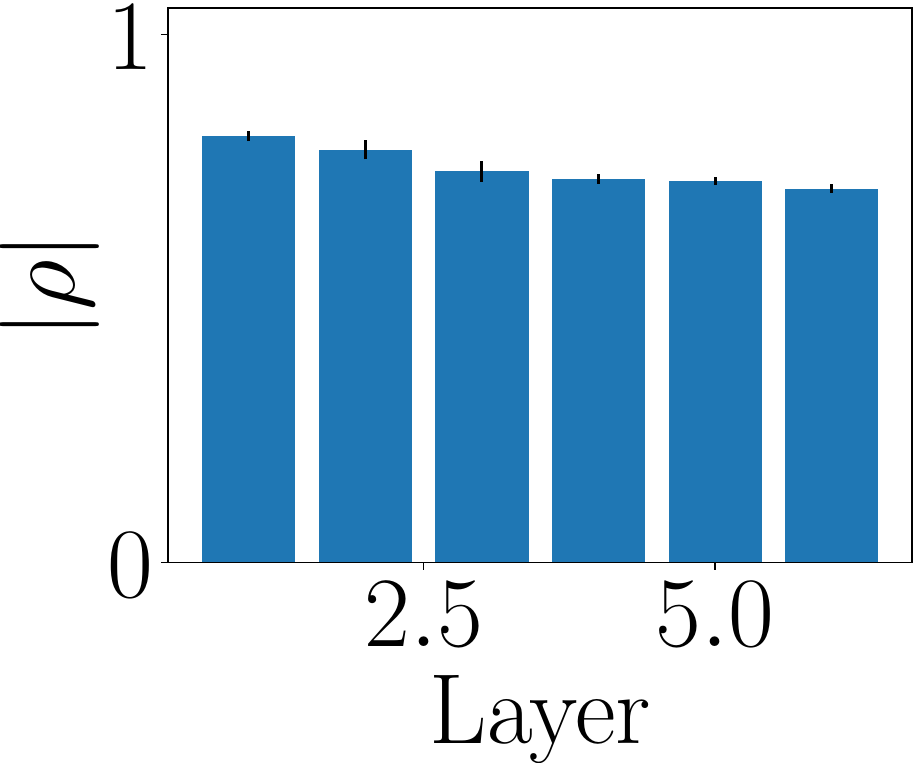}\label{C10La}}
    \subfloat[][ListOps]{\includegraphics[width=0.33\linewidth]{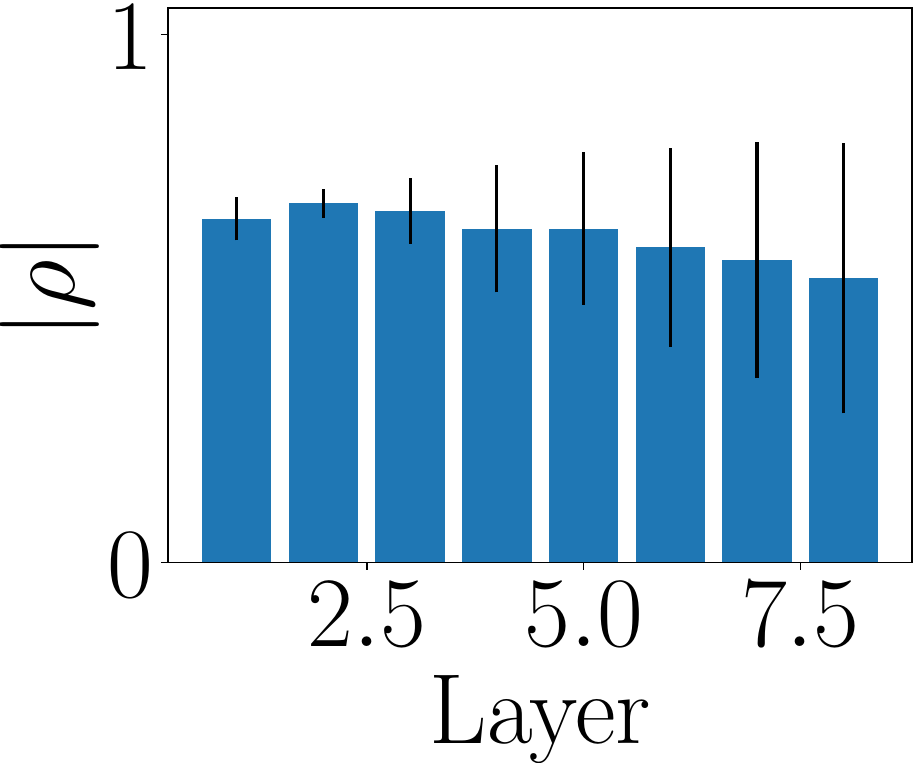}\label{ListOpsLa}}
    \subfloat[][SC]{\includegraphics[width=0.33\linewidth]{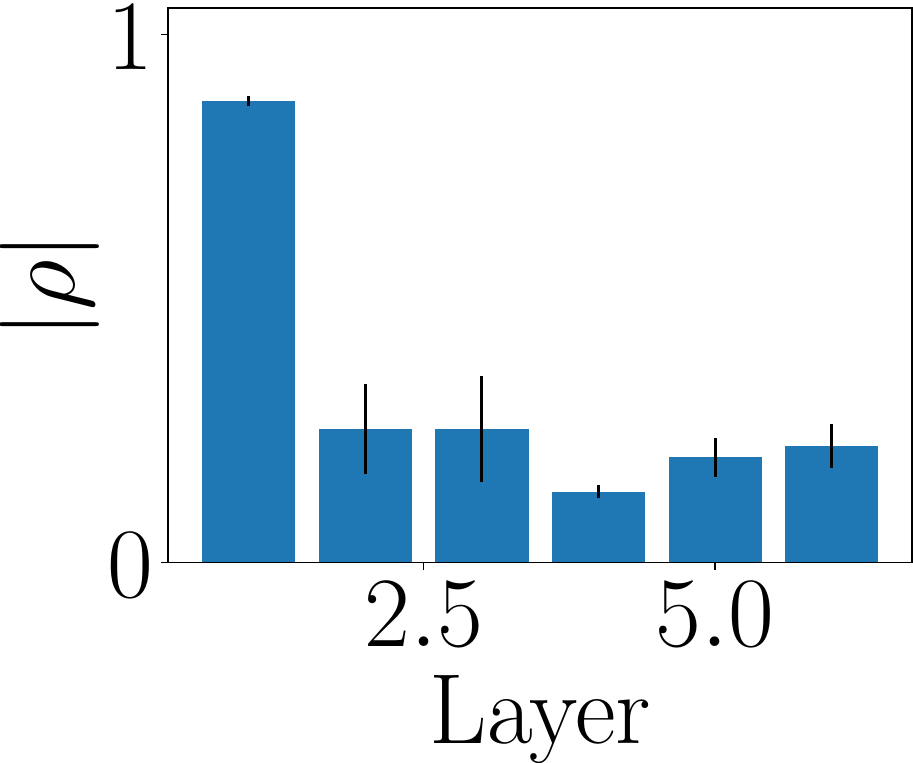}\label{SCLa}}
    \caption{$|\rho|$ of $\bar{R}$ against Layers in training S5 on (i)+(ii).}
    \label{LayerEx}
\end{figure}
\section{K-spectral Metric for Each Layer}
K-spectral metric is averaged over SSMs in our method.
This section gives how K-spectral metric is different among layers.
\rfig{LayerEx} plots absolute values of correlation coefficients for each layer when using S5 on (i)+(ii).
This figure shows that the early layers tend to have high correlation coefficients.
This might be because the input signals of input-side layers are
affected by the current data sequence more than those of output-side layers.
Whereas $|\rho|$ of SC is 0.44 for $\bar{R}$ on S5 in \rtab{Classification}, 
$|\rho|$ of the first layer is closed to one. 
We will investigate the weighting average of the information of spectrum over layers in our future work.

\end{document}